\documentclass[times,twocolumn,final,number]{elsarticle}

\usepackage{ycviu_clean}
\usepackage{framed,multirow}
\usepackage{amsmath}
\usepackage{amssymb}
\usepackage{booktabs}
\usepackage{makecell}
\usepackage{multirow}
\usepackage{graphicx}
\usepackage{caption}
\usepackage[vlined,ruled,commentsnumbered,linesnumbered]{algorithm2e}
\usepackage[colorlinks=true,linkcolor=black,citecolor=black,urlcolor=blue]{hyperref}
\usepackage[separate-uncertainty = true,multi-part-units = repeat]{siunitx}
\usepackage{mathtools}
\usepackage{rotating}

\usepackage{url}
\usepackage{xcolor}
\definecolor{newcolor}{rgb}{.8,.349,.1}

\journal{Computer Vision and Image Understanding}

\begin{document}















\begin{frontmatter}

\title{Uncertainty-Aware Consistency Regularization for Cross-Domain Semantic Segmentation}

\author[1]{Qianyu Zhou \fnref{*}} 
\author[1]{Zhengyang Feng \fnref{*}}
\author[1]{Qiqi Gu}
\author[2]{Guangliang Cheng}
\author[3]{Xuequan Lu\corref{cor1}}
\author[2]{Jianping Shi}
\author[1]{Lizhuang Ma\corref{cor1}}

\address[1]{Shanghai Jiao Tong University, 800 Dongchuan Road, Shanghai and 200240, China}
\address[2]{SenseTime Research, 1900 Hongmei Road, Shanghai and 200233, China}
\address[3]{Deakin University, 75 Pigdons Rd, Waurn Ponds, VIC 3216, Australia}

\fntext[*]{Equal contributions.}
\cortext[cor1]{Joint corresponding author. 
ma-lz@cs.sjtu.edu.cn (Lizhuang Ma) ;
xuequan.lu@deakin.edu.au (Xuequan Lu);}

\begin{abstract}
Unsupervised domain adaptation (UDA) aims to adapt existing models of the source domain to a new target domain with only unlabeled data. 
Most existing methods suffer from noticeable negative transfer resulting from either the error-prone discriminator network or the unreasonable teacher model. Besides, the local regional consistency in UDA has been largely neglected, and only extracting the global-level pattern information is not powerful enough for feature alignment due to the abuse use of contexts.
To this end, we propose an uncertainty-aware consistency regularization method for cross-domain semantic segmentation. Firstly, we introduce an uncertainty-guided consistency loss with a dynamic weighting scheme by exploiting the latent uncertainty information of the target samples. As such, more meaningful and reliable knowledge from the teacher model can be transferred to the student model. We further reveal the reason why the current consistency regularization is often unstable in minimizing the domain discrepancy. Besides, we design a ClassDrop mask generation algorithm to produce strong class-wise perturbations. Guided by this mask, we propose a ClassOut strategy to realize effective regional consistency in a fine-grained manner. 
Experiments demonstrate that our method outperforms the state-of-the-art methods on four domain adaptation benchmarks, \emph{i.e.,} GTAV $\rightarrow $ Cityscapes and SYNTHIA $\rightarrow $ Cityscapes, Virtual KITTI $\longrightarrow$ KITTI and Cityscapes $\longrightarrow$ KITTI.
\end{abstract}


\end{frontmatter}

\section{Introduction}
Semantic segmentation aims to identify the semantic category of each pixel in a given image. 
Recent studies have shown rapid progress with a variety of CNN-based algorithms trained on a large-scale annotated dataset to tackle this problem~\cite{fcn,chen2018deeplab,chen2018encoder,EMANet}. However, due to the time-consuming process of annotating pixel-wise labels~\cite{cordts2016cityscapes}, building such a large annotated dataset is cost-expensive. Compared with manual annotation, the label of synthetic data is much easier to obtain, and thus it is natural to use synthetic data to supervise the segmentation model instead of real data~\cite{gtav,synthia}. However, there always exists a significant performance drop when the learned source models are directly applied to target data, due to the existence of a domain gap between the synthetic images and real images.

To address this issue, various unsupervised domain adaptation (UDA) techniques have been proposed from the domain distribution shift perspective to align the latent feature distributions between the source domain and target domain. Many researchers have exploited additional supervised signals based on the adversarial framework such as depth~\cite{lee2018spigan,GIO_Ada,DADA}, style~\cite{MADAN_NIPS2019,Cycada,Conservative_loss}, category constraint~\cite{FCN_wild,CrossCity}, decision boundary~\cite{mcd,SWD} and other domain-invariant information~\cite{SIBAN} to promote the feature alignment. However, due to the fact that it always requires a domain classifier (discriminator) during the training procedure, these adversarial-based approaches often suffer from training instability and the phenomenon of negative transfer~\cite{CLAN,choi2019self}. 

Consistency regularization is one of the non-adversarial methods exploited in cross-domain segmentation to cope with the negative effect caused by adversarial training~\cite{choi2019self,SEANet}.
This kind of consistency-based methods usually perform the feature-level domain alignment between a student model and a teacher model.
The teacher model is an exponential moving average (EMA) of the student model, and then the teacher model could transfer the learned knowledge to the student. The target predictions of the student and teacher model under different perturbations are penalized by a consistency constraint. 

\begin{figure}
\centering
\includegraphics[scale=0.6]{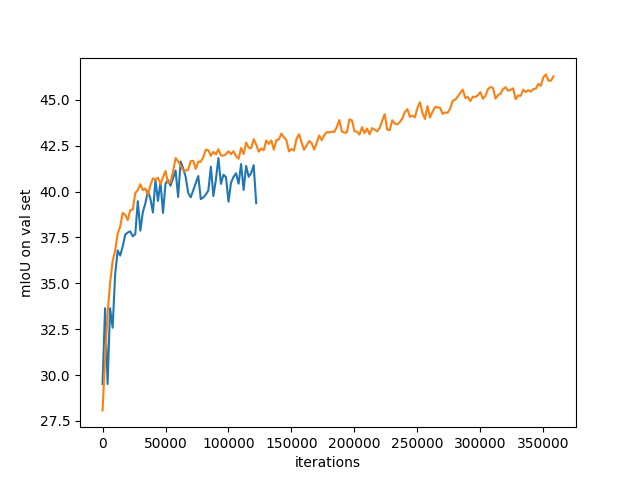}
\caption{ mIoU comparison on the validation set of Cityscapes
by adapting from GTA5 dataset to Cityscapes dataset. The blue
line corresponds to the conventional Mean Teacher strategy
\cite{choi2019self}. The orange line corresponds to the consistency-based adaptation combined with our proposed uncertainty guided module. 
}
\label{fig:comparison}
\end{figure}

In the previous consistency-based works~\cite{choi2019self,medical_self_emsembling}, a common consistency loss, \emph{i.e., } Mean Square Error, is used to ensure the consistency between the student's prediction and the teacher's prediction. We observe that such a simple consistency constraint is usually weak for domain adaptive semantic segmentation, which is reflected in two respects. Firstly, this kind of alignment did not consider the reliability of the teacher predictions, and not all pixel-wise predictions are highly confident for knowledge transfer. Directly imposing a consistency constraint onto all pixels is inappropriate, which could harm the learning process by generating unreasonable guidance for the student model. Secondly, although the whole training of consistency-based adaptation is more stable than adversarial-based adaptation, it is still insufficient. Due to the fact that the basic Mean-Teacher structure may 
trigger the ``error accumulation'', it could take more training iterations to converge and even may lead to early performance degradation during the adaptation process.  The performance curve on the target domain images is shown in Fig. \ref{fig:comparison}.

In the existing consistency regularization methods, \emph{e.g.,} ~\cite{choi2019self,medical_self_emsembling}, the inconsistent penalty is usually adopted on the global level for prediction map, while the region-wise consistency on the local level is ignored, \emph{i.e.,} some contextual object occurrence should be consistent wherever the environments are. Only extracting the global-level pattern information is not powerful enough for the feature-level representation alignment. Without this alignment, the performance will drop significantly in the target domain.  
We attempt to learn the robust representations to varying environments by exploring the fine-grained regional consistency, to prevent the model from abusing the contexts.

Motivated by the above facts, we propose a novel uncertainty-aware consistency regularization scheme to address the domain shift for cross-domain segmentation. Firstly, we introduce a dynamic weighting scheme with an uncertainty-guided consistency loss to capture the understanding of hidden epistemic uncertainty of target predictions for UDA in semantic segmentation. Secondly, we design a ClassDrop mask generation algorithm to produce strong class-wise perturbations. Guided by this mask,  we present an innovative ClassOut strategy to keep the local regional consistency in a fine-grained manner. The whole architecture includes a student model, a teacher model, and our proposed uncertainty module.

In detail, our uncertainty-guided consistency constraints are imposed between the Mean-Teacher system and our proposed uncertainty module, which motivates both the student model and teacher model to alternately promote each other by providing positive feedback, thus leading to the domain gap to be gradually reduced.  To cope with the instability of the conventional consistency regularization framework, we introduce a dynamic weighting scheme of the consistency loss, which is to calculate a time-dependent threshold for filtering out the unreasonable predictions along with mining the highly confident pixel-wise predictions of the target sample. In this manner, the adaptation is realized in a more accurate direction, instead of the rough distribution matching. 
To address the issue of local regional consistency in UDA, we propose a ClassOut strategy to learn more robust region-wise features under varying environments. Our main idea is that the same input image should be invariant under the perturbations by randomly dropping some categories. We design a ClassDrop mask generation algorithm to generate such strong class-wise perturbations. This mask is utilized to filter out the regions of the input target image and the uncertainty mask at the same time to ensure regional consistency on the local level.

Our main contributions are summarized as follows.
\begin{itemize}
    \item We propose an uncertainty-aware consistency regularization framework for cross-domain semantic segmentation, which is a practical, intuitive and elegant contribution to the field. It is also a simple yet effective method for UDA in semantic segmentation.
    
    \item We design an uncertainty-guided consistency loss with a dynamic time-dependent weighting scheme and further reveal the reason why the current consistency regularization is often unstable in minimizing the domain discrepancy. We also show that our method can effectively ease this issue by mining the most reliable and meaningful samples between the source and the target domains.
    
    \item We develop a ClassOut strategy for keeping the local regional consistency in UDA. Meanwhile, we propose a ClassDrop mask generation algorithm to produce strong class-wise perturbations for guiding the ClassOut.
    
    \item We provide extensive experimental results with two common backbone networks, \emph{i.e.,} VGG16 and ResNet101 and show that our approach achieves outstanding performance on four challenging benchmark datasets including both the synthetic-to-real adaptation and cross-city adaptation, \emph{i.e.,} GTAV $\longrightarrow$ Cityscapes, SYNTHIA $\longrightarrow$ Cityscapes, Virtual KITTI $\longrightarrow$ KITTI and Cityscapes $\longrightarrow$ KITTI.
\end{itemize}

\section{Related Work}

\subsection{Semantic Segmentation}

Semantic segmentation is a highly active research field in computer vision. Traditional works of semantic segmentation mainly focused on manually designed image features. With the recent surge of deep learning, a lot of CNN-based methods have been studied and we have witnessed a rapid boost in semantic segmentation performance. Long \emph{et al.}~\cite{fcn} firstly formulated semantic segmentation as a per-pixel classification problem and proposed a fully convolutional network (FCN). With modifications for pixel-wise
prediction, many recent approaches have been proposed, such as DeepLab v2~\cite{chen2018deeplab}, DeepLab v3+~\cite{chen2018encoder}, EMANet~\cite{EMANet}, \emph{etc.} Such models are generally trained on datasets
with pixel-wise annotation, \emph{e.g.,} Cityscapes~\cite{cordts2016cityscapes}, PASCAL~\cite{Pascal} and COCO~\cite{COCO}. However, building such large-scale datasets with dense annotations costs expensive human labor. An alternative approach is to train a model on synthetic
data generated from virtual 3D environments, for example, GTAV~\cite{gtav}, SYNTHIA~\cite{synthia}, \emph{etc.} Unfortunately, when directly applying the model trained on the synthetic data to the real-world scenarios, the performance will be noticeably degraded. The main reason lies in the large domain gap or distribution shift
between the source domain and target domains.

\subsection{Domain Adaptation}
In conventional machine learning, there holds a basic assumption that the training data and testing data are sampled independently from an identical distribution~(\emph{i.i.d}), while this assumption does not always hold in real-world scenarios. Domain Adaptation aims to mitigate the performance drop caused by the distribution mismatch between
training and testing data when applying the trained model into the testing data. Unsupervised Domain Adaptation~(UDA) refers to the setting when the labeled target data is not available. This question has been well studied in image classification. Please refer to \cite{csurka2017domain} for a comprehensive survey.
Conventional methods aim to learn domain-invariant representations
through Maximum Mean Discrepancy (MMD) ~\cite{MMD,long2015learning,cariucci2017autodial,sun2016return}, geodesic flow kernel~\cite{flow_kernel}, sub-space alignment~\cite{sub_space}, asymmetric metric learning\cite{asymmetric_kernel}. Inspired by GAN~\cite{GAN}, adversarial learning is successfully applied in UDA to align the feature distributions from different domains. DANN~\cite{DANN} was the pioneering work, it encouraged a generator to enforce the two distributions to be as close as possible, and to fool the domain classifier at the same time. Most of these UDA methods work on simple and small classification datasets (e.g., MNIST~\cite{MNIST} and SVHN~\cite{SVHN}), and may have limited performance in
more challenging tasks, like semantic segmentation.

\subsection{Domain Adaptation for Semantic Segmentation}
Recently many approaches have been proposed to address the domain shift in semantic segmentation. Pioneered by~\cite{FCN_wild}, Hoffman \emph{et al}.  proposed a domain-adversarial training method by aligning the features between two domains. Following this line, many works have been introduced to address the cross-domain semantic segmentation via the adversarial-based methods, which have achieved great successes in this field. This kind of distribution alignment could be performed at different representation layer, such as pixel-level \cite{Cycada,LSD,chen2019crdoco,DLOW,PCEDA,FDA}, feature level \cite{CrossCity,SIBAN,FCN_wild,CDA,ROAD,DISE,IntraDA} and output level \cite{AdaptSegNet,tsai2019domain,GIO_Ada,CLAN,APODA,SIM}. Many researchers have exploited additional supervised signal based on the adversarial framework such as depth \cite{lee2018spigan,GIO_Ada,DADA}, style ~\cite{MADAN_NIPS2019,Cycada,Conservative_loss}, category constraint ~\cite{FCN_wild,CrossCity}, decision boundary~\cite{mcd,SWD}, and other domain-invariant information to promote the feature alignment. Despite their efforts, these approaches need to maintain an extra discriminator network, thus suffering from training instability and negative transfer~\cite{CLAN,choi2019self}.

To tackle these issues, another line of non-adversarial methods, \emph{e.g.,} self-training~\cite{CBST,CRST,BDL,feng2020semiv1} have been recently studied and applied in the field of UDA.
However, these methods need to generate pseudo labels and fine-tune the segmentation model iteratively in many stages, they cannot be trained end-to-end.
Different from the above self-training approaches, consistency-based methods~\cite{choi2019self,medical_self_emsembling} is a completely different way and a simple online method to learn domain-invariant information in an end-to-end manner. 

\subsection{Consistency Regularization}

Consistency Regularization is applied in the field of semi-supervised learning, which employs unlabeled data to produce consistent
predictions under different perturbations~\cite{Mean_teacher}. Tarvainen \emph{et al.}~\cite{Mean_teacher} firstly encouraged consistency between the predictions of a student network and a teacher network. The teacher's weights are an exponential moving average of those of the student, leading to faster convergence and improved results. French \emph{et al.} \cite{french2018self} then applied the Mean-Teacher framework to the unsupervised domain adaptation for image classification. To address the domain shift for magnetic resonance imaging (MRI), Perone \emph{et al.}~\cite{medical_self_emsembling} applied the self-ensembling method to the medical imaging segmentation task. Considering the UDA task for urban scenes, Choi \emph{et al.}~\cite{choi2019self} proposed a self-ensembling with the GAN-based data augmentation method for cross-domain segmentation. Our work is mostly related to ~\cite{choi2019self}. Inspire by the work \cite{yu2019uncertainty} designed for semi-supervised 3D left atrium segmentation, we propose to capture the latent uncertainty understanding of the teacher model, and encourage the student model to learn from 
that reliable knowledge.

\begin{figure*} 
\centering
\includegraphics[scale=1.5]{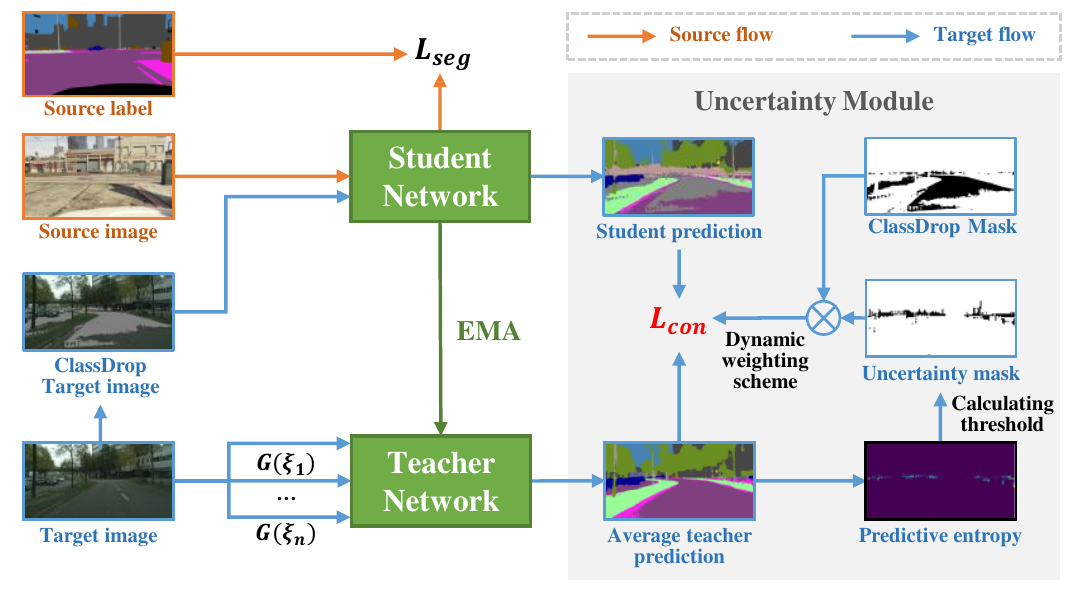}
\caption{ An overview of the proposed framework. The whole framework includes a student network, a teacher network updated by exponential moving average (EMA), and our uncertainty module. A ClassDrop Mask is generated by the target image and then used to filter the local regions of the target image. The ClassDrop target image is fed into the student network to get student prediction.
We employ different augmentations $G(\xi_{i})$ for the input target sample, and they are fed into the teacher model. In our uncertainty module, we perform $N$ times stochastic forward passes to get an average teacher prediction. Then, with the estimation of predictive entropy and the proposed dynamic threshold, we could get the uncertainty mask.
Thus, the ClassDrop Mask is element-wise multiplied with the uncertainty mask for filtering out the unreasonable predictions. Guided by the proposed dynamic weighting scheme and ClassOut strategy, our uncertainty-guided consistency loss $L_{con}$ could encourage the teacher model to transfer more reliable knowledge to the student.}
\label{fig:framework}
\end{figure*}

\section{Methodology}
In this section, we present our uncertainty-aware consistency regularization method for unsupervised domain adaptive segmentation. Following the unsupervised domain adaptation protocol~\cite{CrossCity,FCN_wild,ROAD}, the synthetic data is utilized as the source domain $S$, and the real data as target domain $T$. In the source domain, we have access to the synthetic images $x_s \in S$ along with their corresponding ground-truth labels $y_s$. In
the target domain, only unlabeled images $x_t \in T$ are available.

\subsection{Overview}
The overview of our proposed uncertainty-aware consistency regularization method is illustrated in Fig.~\ref{fig:framework}. 
The whole framework includes three modules: a student model $f_S$, a teacher model $f_T$, and our uncertainty module. The key idea is to decrease the uncertainty of the error-prone teacher model as training progress thus leading the adaptation process in a more accurate and stable way. 

Specifically, a ClassDrop mask is generated by the target image to provide strong class-wise perturbations by randomly dropping some classes that are presented in the target image. This mask will be utilized to filter out the local regions of both the target image and the uncertainty mask (a mask we defined to indicate the uncertain pixels). For the former, a ClassDrop target image is fed into the student network to get student prediction. For the latter, we will explain the data flow in detail. Firstly, we employ data augmentation, \emph{e.g.,} Gaussian Noise, for the input target samples. In our proposed uncertainty module, we perform stochastic forward passes to calculate the mean of target predictions. In this way, we are able to employ our teacher model as a Bayesian network to estimate the latent uncertainty information of the teacher predictions. We formulate the uncertainty as the pixel-wise predictive entropy. Then, we calculate a time-dependent threshold for filtering out those unreasonable predictions along with mining the high confident pixel-wise predictions of the target sample. Thus, the ClassDrop Mask is element-wise multiplied with the uncertainty mask for filtering out the unreasonable predictions. With the help of the proposed dynamic weighting scheme, an uncertainty-guided consistency loss is penalized to target predictions under different perturbations, which could lead the student model to gradually learn from the more meaningful and reliable predictions of the teacher model during the training process. 

\begin{figure} 
\centering
\includegraphics[scale=0.8]{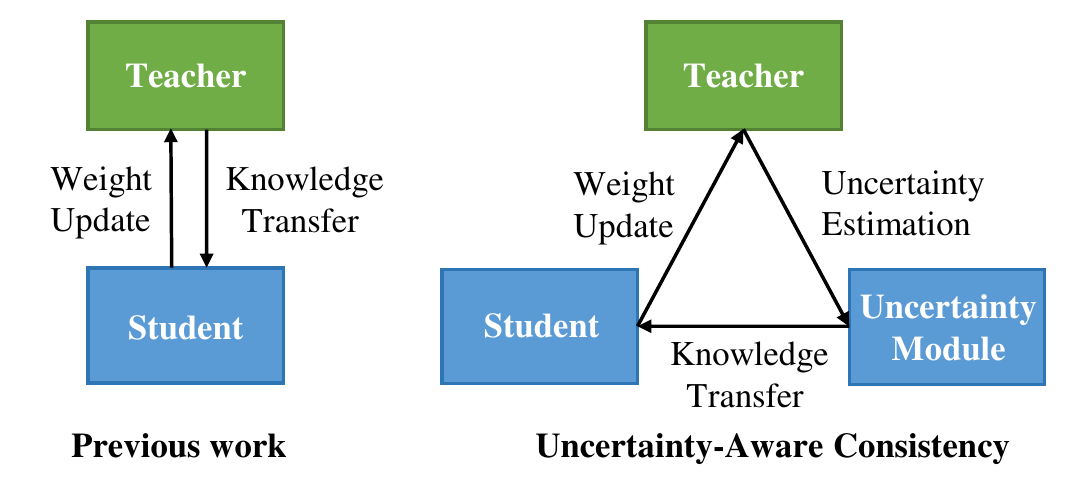}
\caption{ Previous work vs. our approach.}
\label{fig:triangular}
\end{figure}

\subsection{Uncertainty Module}

As shown in Fig.~\ref{fig:triangular}, the uncertainty module serves as a bridge for connecting the teacher model and the student model. According to the uncertainty estimation method in Bayesian
networks~\cite{kendall2017uncertainties}, we are motivated to capture the understanding of epistemic uncertainty using stochastic forward passes. In step 1, we perform the stochastic forward pass and then extract the uncertainty information of the error-prone teacher model. We formulate the uncertainty as the pixel-wise predictive entropy. In step 2, we calculate the uncertainty mask given our time-dependent threshold, and a ClassDrop mask given the target image. Guided by these masks, we enforce an uncertain-aware consistency loss and a ClassOut Strategy onto the student predictions and the teacher predictions, thus the student model could learn credible knowledge from the teacher.

The teacher's weights $\Phi^{'}_{t}$ at training step $t$ are updated by the student's weights $\Phi_{t}$  with a smoothing coefficient $\alpha \in [0,1]$, which can be formulated as follows:
\begin{align}
  \Phi^{'}_{t} & = \alpha \cdot \Phi^{'}_{t-1} + (1-\alpha) \cdot \Phi_{t} , 
\end{align}
where $\alpha$ refers to the EMA decay that controls the updating rate.

 Specifically, we make $N$ copies of the target image and inject a Gaussian noise for the target predictions following prior works~\cite{choi2019self}. Then, we perform $N$ stochastic forward passes for the target teacher sample to get the average teacher prediction.
Given a set of pixel-wise predicted class scores $\{\boldsymbol{P}_{i}^{(h, w, c)}(x_t)\}_{i=1}^{N}$ of the target samples, the average teacher prediction is formulated as:
\begin{align}
  \hat{P_{c}} & = \frac{1}{N}\sum\limits_{i=1}^{N}\boldsymbol{P}_{i}^{(h, w, c)}(x_t),
\end{align}
where $\hat{P_{c}}$ denotes the mean of the predictive probability of the $c$-th class after $N$ times stochastic forward passes. Thus, the pixel-wise predictive entropy is as follows: 

\begin{align}
  \label{eq:entropy}
  \mu^{(h,w)} & =-\sum\limits_{c=1}^{C}\hat{P_{c}}\cdot log(\hat{P_{c}}),
\end{align}
where $\zeta$ refers to the predictive entropy in pixel level. All the volumes of each pixel's uncertainty forms a set $Z=\{\zeta\}_{i=1}^{N}$.

\begin{figure} 
\centering
\includegraphics[scale=1.5]{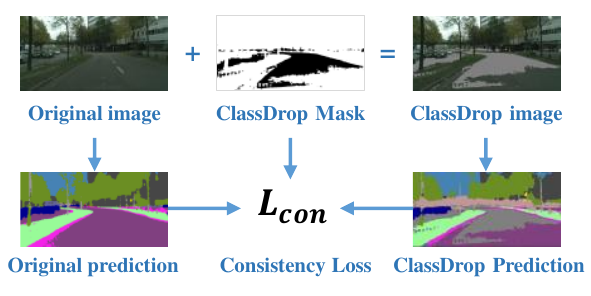}
\caption{The main idea of the ClassOut strategy is that the same input image should be invariant under the ClassDrop perturbations. Specifically, we firstly generate a ClassDrop mask from an original target image. Then, guided by this mask, we calculate a consistency loss between the original prediction from the teacher and the ClassDrop prediction from the student. Therefore, we can keep the local regional consistency on a fine-grained level.}
\label{fig:motivation}
\end{figure}

\subsection{Dynamic Weighting Scheme}
\label{section:uncertainty}
With the help of the uncertainty of each pixel, we could calculate a dynamic threshold to filter out the unreliable pixel-wise prediction.
On top of that, certain pixels with high confident probabilities will be left and the student model could gradually learn the reliable target predictions from the teacher model. 

In particular, we first calculate the uncertainty threshold $R$ to select the confident pixels according to the uncertainty map we have estimated. 
Inspired by the ramp-up function of consistency weight in \cite{choi2019self}, we came up with the Eq. \ref{eq:thres}, which dynamically increases the threshold while the uncertainty is decreased during the training process. This design is a time-dependent ramp-up function, which changes dynamically over time:
\begin{equation}
    \label{eq:thres}
    R=\alpha+(1-\alpha) \cdot e^{\beta (1-t/t_{max})^2} \cdot Z_{sup} 
\end{equation}
where $t$ denotes the current training step and $t_{max}$ is the  maximum training step. $Z_{sup}$ means the upper bound of the volumes' self-information, which is denoted by $Z_{sup}=sup\{\mu\}_{i=1}^{N}$. And $\alpha$ and $\beta$ are two hyper-parameters. 

 The uncertainty-aware consistency loss $L_{con}$ is imposed between the prediction maps extracted from the student and the predictions from the teacher network. 

\begin{equation}
\label{consis}
    \begin{aligned}
L_{con}(f_S,f_T)& = 
        \sum_{h=1}^{H} \sum_{w=1}^{W} I(\mu^{(h, w)}< R)\\
        &~~~~\cdot  ||f_{\theta(x_{T_1})^{(h, w)}}) - f_{\theta^{\prime}}(x_{T_2})^{(h, w)})||^{2},
\end{aligned}
\end{equation}
where $I$ is an indicator function,  and $x_{T_1}$ and $x_{T_2}$ are two input target samples with different augmentations.  $f_{\theta(x_{T})}$ and $f_{\theta^{\prime}}(x_{T})$ are the student and teacher prediction map after the softmax function, respectively.
Note that the prediction map  $f_{\theta^{\prime}}(x_{T_2})$ used for consistency regularization is the stochastic one rather than the average one. Our uncertainty mask $M_{uncertainty}=I(\mu^{(h, w)}< R)$ can  reweigh not only the Mean Squared Error (MSE) loss
but also the Cross-Entropy Loss. For simplicity, we use the Mean Squared Error (MSE) in this paper.

\subsection{ClassOut Strategy}
\label{section:classout}

Previous consistency regularization methods, \emph{e.g.,} ~\cite{choi2019self,medical_self_emsembling}, usually impose the inconsistent penalty on the global level for prediction map, while the region-wise consistency on the local level is largely ignored, \emph{i.e.,} some contextual object occurrence should be consistent whatever the environments are. Only extracting the global-level pattern information is not powerful enough for the feature-level representation alignment. Due to the lack of local regional consistency, the performance will drop significantly in the target domain.  
Our goal is to learn the robust representations to varying environments by exploring the fine-grained regional consistency, to prevent the model from abusing the contexts.

Firstly, we propose an innovative ClassDrop mask generation algorithm to provide strong class-wise perturbations, as shown in Algorithm \ref{algorithm 0}. To be specific, we firstly get the pseudo labels $\tilde{Y}_{T}$ from the target predictions $f_{\theta^{\prime}}(X_T)$. The set of the classes presented in $\tilde{Y}_{T}$ are noted as $C$. 
We get a class ratio $\delta$ sampled from a uniform distribution. Then, we randomly select $\delta |C| $ classes in $C$. A binary mask M is
generated by setting the pixels from
those classes to 1 in M, whereas all others will have a value 0. This mask is utilized to filter out the local regions in both the target image and the uncertainty mask by an element-wise multiplication. 

\begin{algorithm}
	\caption{ClassDrop Mask Generation Algorithm}\label{algorithm 0}
     \KwIn{teacher model $f_{\theta^{\prime}}$, target image $X_T$, min class ratio $a$, and max class ratio $b$.}
      \KwOut{ClassDrop mask $M$}
      $\hat{f}_{\theta^{\prime}} \leftarrow  f_{\theta^{\prime}}(X_T)$\;
      $\tilde{Y}_{T} \leftarrow \arg \max _{c^{\prime}}\; \hat{f}_{\theta^{\prime}} \left(i, j, c^{\prime}\right)$\;
      $C \leftarrow$ Set of the classes present in $\tilde{Y}_{T}$\;
      $\delta \leftarrow U(a, b)$ \;
      $c \leftarrow$ Randomly select $\delta |C| $ classes in $C$\;
     \For{each $i,j$ }{
        $M(i, j)=\left\{\begin{array}{l}1, \text { if } \tilde{Y}_{T}(i, j) \in c \\ 0, \text { otherwise }\end{array}\right.$\
      }
      
      
     return $M$\;
\end{algorithm}
The main idea of the ClassOut strategy is that the same input image should be invariant under the ClassDrop perturbations. Thus, guided by the ClassDrop mask, we calculate a consistency loss between the original prediction from the teacher and the ClassDrop prediction from the student. Therefore, we can keep the local regional consistency:
\begin{align}
\label{classout}
 L_{\text {con}}=\left\|M \odot\left(f_{\theta^{\prime}}(M \odot X_T)-f_{\theta^{\prime}}(X_T)\right)\right\|^{2},
\end{align}

\subsection{Unified Training}
\noindent \textbf{Consistency Loss:} By unifying the ClassOut strategy and the dynamic weighting scheme into the same framework to realize end-to-end training, we reformulate the consistency loss as follows:
\begin{align}
\label{classout}
 L_{\text {con}}=\left\|M_{classdrop} \odot M_{uncertainty} \odot \left(f_{\theta}(M_{classout} \odot X_T)-f_{\theta^{\prime}}(X_T)\right)\right\|^{2},
\end{align}
where the final consistency loss is reweighted by the uncertainty mask $M_{uncertainty}$ (defined in section \ref{section:uncertainty}) and the ClassDrop mask $M_{classout}$ (defined in section \ref{section:classout}). In other words, the reweighted mask of the consistency loss is the element-wise multiplication between the uncertainty mask $M_{uncertainty}$ and the ClassDrop mask $M_{classout}$.
We simplify the previous definition and reformulate it as follows:
\begin{equation}M_{classdrop}(i, j)=\left\{\begin{array}{l}
1, \text { if } \tilde{Y}_{T}(i, j) \in c_{remain} \\
0, \text { otherwise }
\end{array}\right.\end{equation}

\begin{equation}M_{uncertainty}(i, j)=\left\{\begin{array}{l}
1, \text { if } \mu(i, j)< R \\
0, \text { otherwise }
\end{array}\right.\end{equation}
where $c_{remain}=\delta |C|$ is the selected classes from a class set, $\mu$ is the predicted entropy defined in Eq. \ref{eq:entropy} and $R$ is the dynamic threshold defined in Eq. \ref{eq:thres}. 

\noindent \textbf{Supervised Loss:} The segmentation loss $L_{seg}$ is the cross-entropy loss for optimizing the images from the source domain, which can be defined as:
\begin{align}
  L_{seg} & = -\sum\limits_{h=1}^{H}\sum\limits_{w=1}^{W} \sum\limits_{c=1}^{C} y_{s}^{(h,w,c)}log(P_s^{(h,w,c)}),
\end{align}
where $y_{s}$ is the ground truth for source images and $P_s = f_{S}((\hat{x}_s)^{(h, w, c)})$ is the segmentation output of source-translated input images.

\noindent \textbf{Total Loss:} The total loss $L_{total}$ is the weighted sum of the segmentation loss $L_{seg}$ and the consistency loss $L_{con}$, and can be written as:
\begin{align}
  \label{equa:totoal}
  L_{total} & = L_{seg}+\lambda_{con}L_{con},
\end{align}
where $L_{con}$ is the combination of Equation \ref{consis}
and Equation \ref{classout}.
 $\lambda_{con}$ is the dynamic weight of the consistency loss. To balance the segmentation loss and the consistency loss, we use the same ramp-up function $\lambda_{con}$ as \cite{tranheden2020dacs}. It is to increase the dominance of $L_{seg}$ during the early training steps and to increase the dominance of  $L_{con}$ during the late training steps.

\subsection{Discussion}
In this subsection, we will discuss the main differences between the existing research and our proposed method.

There exist some works \cite{wen2019bayesian,kurmi2019attending} which used uncertainty estimation in domain adaptation; however, those methods \cite{wen2019bayesian,kurmi2019attending} always need to maintain a Bayesian Discriminator in adversarial training, thus suffering the drawbacks of negative transfer and remarkable instability of training. Besides, their methods only work well on the simple and small classification dataset, and can hardly work well in structured tasks, \emph{e.g.,} semantic segmentation. Therefore, we do not compare the experimental results with these methods in Section 4. Our uncertainty-aware consistency regularization shows that a non-adversarial approach can achieve the state-of-the-art as well without the need of maintaining an extra discriminator network or carefully tuning the optimization procedure for min-max problems during the domain adaptation procedure. 

Different from \cite{han2019unsupervised,zheng2020unsupervised}, we focus on investigating the problem of ``error accumulation'' in consistency regularization, rather than self-training. In contrast to \cite{yu2019uncertainty} that targets the semi-supervised learning for the 3d left atrium segmentation task, while we target the unsupervised domain adaptation for the image semantic segmentation task. \textit{Our method differs from these approaches in  several aspects. } Firstly, we propose a dynamic weighting scheme and a ClassOut strategy for the uncertainty-consistency loss. The uncertainty mask, Classdrop mask are employed in a completely different way from previous works \cite{wen2019bayesian,kurmi2019attending,han2019unsupervised, yu2019uncertainty}. We further reveal the reason why the current consistency regularization is often unstable in minimizing the distribution discrepancy in Section 1 and Section 4.1. Besides, we also show that our method can effectively ease this issue by mining the most reliable and meaningful samples between the source and the target domains. To sum up, our uncertainty-aware consistency regularization framework is a practical, intuitive and elegant contribution to the field, and it is also a simple yet effective unsupervised domain adaptation method for semantic segmentation. To our best knowledge, there are no such domain adaptive segmentation methods published before.

\section{Experiments}
 
In this section, we verify the effectiveness of our method with two common backbone networks, \emph{i.e.,} VGG16 and ResNet 101, on both the synthetic-to-real adaptation and cross-city adaptation on four challenging benchmark datasets, \emph{i.e.,} GTAV $\longrightarrow$ Cityscapes, SYNTHIA $\longrightarrow$ Cityscapes, Virtual KITTI $\longrightarrow$ KITTI and  Cityscapes $\longrightarrow$ KITTI.
\begin{table*}
\caption{Comparison results (mIoU) from GTAV to Cityscapes (with VGG16 backbone). }
\label{table:gtav}
\centering
\resizebox{\textwidth}{!}{%
\begin{tabular}{c|ccccccccccccccccccc|c}
\toprule
Method                 & \begin{turn}{90}road\end{turn} & \begin{turn}{90}sidewalk\end{turn} & \begin{turn}{90}building\end{turn} & \begin{turn}{90}wall\end{turn} & \begin{turn}{90}fence\end{turn} & \begin{turn}{90}pole\end{turn} & \begin{turn}{90}light\end{turn} & \begin{turn}{90}sign\end{turn} & \begin{turn}{90}vegetation\end{turn} & \begin{turn}{90}terrain\end{turn} & \begin{turn}{90}sky\end{turn} & \begin{turn}{90}person\end{turn} & \begin{turn}{90}rider\end{turn} & \begin{turn}{90}car\end{turn} & \begin{turn}{90}truck\end{turn} & \begin{turn}{90}bus\end{turn} & \begin{turn}{90}train\end{turn} & \begin{turn}{90}motocycle\end{turn} & \begin{turn}{90}bike\end{turn} & \begin{turn}{90}\textbf{mIoU}\end{turn}  \\ 
\midrule
Source Only &61.0 &18.5 &66.2 &18.0 &19.6 &19.1 &22.4 &15.5 &79.6 &28.5 &58.0 &44.5 &1.7& 66.6 &14.1 &1.1 &0.0 &3.2 &0.7 &28.3  \\ 
\midrule
SIBAN \cite{SIBAN} & 83.4 & 13.0 & 77.8 & 20.4 & 17.5 & 24.6 & 22.8 & 9.6 & 81.3 & 29.6 & 77.3 & 42.7 & 10.9 & 76.0 & 22.8 & 17.9 & 5.7 & 14.2 & 2.0 & 34.2 \\
CyDADA  \cite{Cycada}   &85.2& 37.2& 76.5& 21.8& 15.0 &23.8& 22.9& 21.5& 80.5& 31.3& 60.7& 50.5& 9.0& 76.9& 17.1& 28.2& 4.5& 9.8& 0.0& 35.4 \\ 
AdaptSegNet \cite{AdaptSegNet} & 87.3& 29.8& 78.6& 21.1& 18.2& 22.5& 21.5& 11.0& 79.7& 29.6 &71.3& 46.8& 6.5& 80.1& 23.0& 26.9& 0.0& 10.6& 0.3& 35.0  \\ 
ROAD~\cite{ROAD}& 85.4& 31.2& 78.6& 27.9& 22.2& 21.9& 23.7& 11.4& 80.7& 29.3 &68.9& 48.5 &14.1 &78.0 &19.1& 23.8& 9.4 &8.3& 0.0& 35.9 \\ 
CLAN~\cite{CLAN} &88.0 & 30.6 & 79.2 & 23.4 & 20.5 & 26.1 & 23.0 & 14.8 & 81.6 & 34.5 & 72.0 & 45.8 & 7.9 & 80.5 & 26.6 & 29.9 & 0.0 & 10.7 & 0.0 & 36.6\\
AdaptPatch~\cite{tsai2019domain} & 87.3 & 35.7 & 79.5 & 32.0 & 14.5 & 21.5 & 24.8 & 13.7 & 80.4 & 32.0 & 70.5 & 50.5 & 16.9 & 81.0 & 20.8 & 28.1 & 4.1 & 15.5 & 4.1 & 37.5 \\
APODA~\cite{APODA} &88.4 &34.2 &77.6 &23.7 &18.3 &24.8 &24.9 &12.4 &80.7 &30.4 &68.6 &48.9 &17.9 &80.8 &27.0 &27.2 &6.2 &19.1 &10.2 &38.0\\
CrCDA~\cite{CrCDA} &86.8  &37.5 &80.4 &30.7 &18.1 &26.8 &25.3 &15.1 &81.5 &30.9 &72.1 &52.8 &19.0 &82.1 &25.4 &29.2 &10.1 &15.8 &3.7 &39.1 \\
SWD~\cite{SWD}  & 91.0 & 35.7 & 78.0 & 21.6 & 21.7 & 31.8 & 30.2 & 25.2 & 80.2 & 23.9 & 74.1 & 53.1 & 15.8 & 79.3 & 22.1 & 26.5 & 1.5 & 17.2 & 30.4 & 39.9 \\
\midrule
DCAN~\cite{DCAN}  &82.3& 26.7 &77.4& 23.7& 20.5& 20.4& 30.3& 15.9& 80.9& 25.4& 69.5& 52.6& 11.1& 79.6& 24.9& 21.2& 1.3& 17.0& 6.7& 36.2  \\ 
CrDoCo~\cite{chen2019crdoco} & 89.1 & 33.2 & 80.1 & 26.9 & 25.0 & 18.3 & 23.4 & 12.8 & 77.0 & 29.1 & 72.4 & 55.1 & 20.2 & 79.9 & 22.3 & 19.5 & 1.0 & 20.1 & 18.7 & 38.1 \\
CDA~\cite{CDA}  &72.9& 30.0& 74.9& 12.1& 13.2& 15.3& 16.8& 14.1& 79.3& 14.5 &75.5& 35.7& 10.0& 62.1& 20.6 &19.0 &0.0 &19.3& 12.0& 31.4 \\ 

CBST~\cite{DCAN} &66.7 &26.8 &73.7& 14.8& 9.5& 28.3& 25.9& 10.1& 75.5& 15.7& 51.6& 47.2& 6.2& 71.9& 3.7& 2.2& 5.4& 18.9& \textbf{32.4}& 30.9 \\ 
ADVENT~\cite{advent}  & 86.8 & 28.5 & 78.1 & 27.6 & 24.2 & 20.7 & 19.3 & 8.9 & 78.8 & 29.3 & 69.0 & 47.9 & 5.9 & 79.8 & 25.9 & 34.1 & 0.0 & 11.3 & 0.3 & 35.6\\
PyCDA~\cite{PyCDA}  & 86.7 & 24.8 & 80.9 & 21.4 & \textbf{27.3} & 30.2 & 26.6 & 21.1 & 86.6 & 28.9 & 58.8 & 53.2 & 17.9 & 80.4 & 18.8 & 22.4 & 4.1 & 9.7 & 6.2 & 37.2\\
\midrule
LSD-seg~\cite{LSD} & 88.0& 30.5& 78.6& 25.2& 23.5 & 16.7& 23.5& 11.6& 78.7& 27.2& 71.9& 51.3& 19.5& 80.4& 19.8& 18.3& 0.9 &20.8 &18.4 &37.1 \\ 
SSF-DAN \cite{du2019ssf}  & 88.7 & 32.1 & 79.5 & 29.9 & 22.0 & 23.8 & 21.7 & 10.7 & 80.8 & 29.8 & 72.5 & 49.5 & 16.1 & 82.1 & 23.2 & 18.1 & 3.5 & 24.4 & 8.1 & 37.7     \\
Conservative Loss~\cite{Conservative_loss}   & 85.6& 38.3& 78.6 &27.2 &18.4 &25.3 &25.0 &17.1 &81.5& 31.3& 70.6 &50.5& 22.3& 81.3& 25.5 &21.0 &0.1 &18.9& 4.3& 38.1 \\ 
PIT \cite{PIT} &86.2 &35.0 &82.1 &31.1 &22.1 &23.2 &29.4 &28.5 &79.3 &31.8 &81.9 &52.1 &23.2 &80.4 &29.5 &26.9 &30.7 &20.5 &1.2 &41.8  \\ 
BDL \cite{BDL} &89.2 & 40.9 & 81.2 & 29.1 & 19.2 &14.2 & 29.0 & 19.6& 83.7& 35.9 & 80.7 & 54.7 & 23.3 & 82.7 & 25.8 & 28.0 & 2.3 & \textbf{25.7} &19.9 & 41.3  \\ 
SIM \cite{SIM}  &88.1 &35.8 &83.1 &25.8 &23.9 &29.2 &28.8 &28.6 &83.0 &36.7 &82.3 &53.7 &22.8 &82.3 &26.4 &38.6 &0.0 &19.6 &17.1 &42.4  \\ 
TGCF-DA + SE \cite{choi2019self} & 90.2& 51.5& 81.1& 15.0& 10.7& \textbf{37.5}& \textbf{35.2} & 28.9 & 84.1 &32.7 & 75.9& \textbf{62.7}& 19.9 & 82.6& 22.9& 28.3& 0.0& 23.0& 25.4& 42.5 \\    
\midrule
\textbf{Ours}  & \textbf{95.1} & \textbf{66.5} & \textbf{84.7} & \textbf{35.1} & 19.8 & 31.2 & 35.0 & \textbf{32.1} & \textbf{86.2} & \textbf{43.4} &\textbf{82.5} & 61.0 & \textbf{25.1} & \textbf{87.1} & \textbf{35.3} & \textbf{46.1} & 0.0 & 24.6 & 17.5 & \textbf{47.8}  \\  
\bottomrule
\end{tabular}}
\end{table*}

\begin{table*}
\caption{Comparison results (mIoU) from SYNTHIA to Cityscapes (with VGG16 backbone).}
\label{table:synthia}
\centering
\resizebox{\textwidth}{!}{%
\begin{tabular}{c|cccccccccccccccc|c|c} 
\toprule
Method                
& \begin{turn}{90}road\end{turn} & \begin{turn}{90}sidewalk\end{turn} & \begin{turn}{90}building\end{turn} & \begin{turn}{90}wall\end{turn} & \begin{turn}{90}fence\end{turn} & \begin{turn}{90}pole\end{turn} & \begin{turn}{90}light\end{turn} & \begin{turn}{90}sign\end{turn} & \begin{turn}{90}vegetation\end{turn} & \begin{turn}{90}sky\end{turn} & \begin{turn}{90}person\end{turn} & \begin{turn}{90}rider\end{turn} & \begin{turn}{90}car\end{turn} & \begin{turn}{90}bus\end{turn} & \begin{turn}{90}motocycle\end{turn} & \begin{turn}{90}bike\end{turn} & \begin{turn}{90}\textbf{mIoU}\end{turn} & \begin{turn}{90}\textbf{mIoU*}\end{turn}  \\ 
\midrule
Source Only &6.8 &15.4 &56.8 &0.8 &0.1 &14.6 &4.7& 6.8 &72.5& 78.6& 41.0& 7.8& 46.9& 4.7& 1.8 &2.1 &22.6& 24.1  \\ 
\midrule
Cross-city~\cite{CrossCity} & 62.7 & 25.6 &  78.3 & - & - & - &  1.2 &  5.4  & 81.3 &  81.0 &  37.4 &  6.4 &  63.5 &  16.1 &  1.2 &  4.6 &  - & 35.7 \\
SIBAN~\cite{SIBAN} & 70.1 & 25.7 & 80.9 & - & - & - & 3.8 & 7.2 & 72.3 & 80.5 & 43.3 & 5.0 & 73.3 & 16.0 & 1.7 & 3.6 & - & 37.2  \\
ROAD~\cite{ROAD} &77.7& 30.0& 77.5& 9.6& 0.3& 25.8& 10.3& 15.6& 77.6& 79.8& 44.5& 16.6& 67.8& 14.5& 7.0& 23.8 &36.2&-  \\ 
AdaptSegNet~\cite{AdaptSegNet} & 78.9& 29.2& 75.5& -& - &- &0.1& 4.8& 72.6& 76.7& 43.4& 8.8& 71.1& 16.0& 3.6& 8.4& -& 37.6  \\ 
CLAN~\cite{CLAN} & 80.4 & 30.7 & 74.7 & -& - &- &1.4 &8.0 &77.1 &79.0 &46.5 &8.9  &73.8 &18.2 &2.2 &9.9 & - &39.3\\
AdaptPatch~\cite{tsai2019domain} & 72.6 & 29.5 & 77.2 & 3.5 & 0.4 & 21.0 & 1.4 & 7.9 & 73.3 & 79.0 & 45.7 & 14.5 & 69.4 & 19.6 & 7.4 & 16.5 & 33.7 & 39.6 \\
SPIGAN~\cite{lee2018spigan}  & 71.1 & 29.8 & 71.4 & 3.7 & 0.3 & 33.2 & 6.4 & 15.6 & 81.2 & 78.9 & 52.7 & 13.1 & 75.9 & \textbf{25.5} & 10.0 & 20.5 & 36.8 & -\\
CrCDA~\cite{CrCDA} &74.5 &30.5 &78.6 &6.6 &0.7 &21.2 &2.3 &8.4 &77.4 &79.1 &45.9 &16.5 &73.1 &24.1 &9.6 &14.2 &35.2 &41.1 \\
APODA~\cite{APODA} &82.9 &31.4 &72.1 & -& - &- &10.4 &9.7 &75.0 &76.3 &48.5 &15.5 &70.3 &11.3 &1.2 &29.4 &- &41.1 \\
SWD~\cite{SWD} & 83.3 & 35.4 & 82.1 & - & - & - & 12.2 & 12.6 & 83.8 & 76.5 & 47.4 & 12.0 & 71.5 & 17.9 & 1.6 & 29.7 & - & 43.5 \\
\midrule
CrDoCo~\cite{chen2019crdoco} & 62.2 & 21.2 & 72.8 & 4.2 & 0.8 & 30.1 & 4.1 & 10.7 & 76.3 & 73.6 & 45.6 & 14.9 & 69.2 & 14.1 & 12.2 & 23.0 &33.4 & - \\
DCAN~\cite{DCAN}  &79.9& 30.4& 70.8& 1.6& 0.6& 22.3& 6.7& 23.0& 76.9& 73.9& 41.9& 16.7& 61.7& 11.5& \textbf{10.3}& 38.6& 35.4&-  \\ 
CDA~\cite{CDA}  &65.2& 26.1& 74.9& 0.1& 0.5& 10.7& 3.7& 3.0& 76.1& 70.6& 47.1 &8.2& 43.2& 20.7& 0.7& 13.1& 29.0& 34.8  \\ 
CBST~\cite{CBST}&69.6& 28.7& 69.5& \textbf{12.1} & 0.1& 25.4& 11.9& 13.6& 82.0& \textbf{81.9}& 49.1& 14.5& 66.0& 6.6& 3.7& 32.4& 35.4& 36.1  \\
ADVENT~\cite{advent} & 67.9 & 29.4 & 71.9 & 6.3 & 0.3 & 19.9 & 0.6 & 2.6 & 74.9 & 74.9 & 35.4 & 9.6 & 67.8 &21.4 & 4.1 & 15.5 & 31.4 & 36.6 \\
PyCDA~\cite{PyCDA} & 80.6 & 26.6 & 74.5 & 2.0 & 0.1 & 18.1 & \textbf{13.7} & 14.2 & 80.8 & 71.0 & 48.0 & 19.0 & 72.3 & 22.5 & 12.1 & 18.1 & 35.9 & 42.6\\
\midrule
Conservative Loss~\cite{Conservative_loss} & 80.0& 31.4& 72.9& 0.4& 0.0& 22.4& 8.1& 16.7& 74.8& 72.2& 50.9 &12.7 &53.9 &15.6 &1.7 &33.5 &34.2& 40.3   \\ 
LSD-seg~\cite{LSD} &80.1& 29.1& 77.5& 2.8& 0.4& 26.8& 11.1& 18.0& 78.1& 76.7& 48.2& 15.2& 70.5& 17.4& 8.7& 16.7& 36.1&-  \\ 
GIO-Ada~\cite{GIO_Ada}  & 78.3 & 29.2 & 76.9 & 11.4 & 0.3 & 26.5 & 10.8 & 17.2 & 81.7 & \textbf{81.9} & 45.8 & 15.4 & 68.0 & 15.9 & 7.5 & 30.4 & 37.3 & 43.0 \\
SSF-DAN~\cite{du2019ssf}  & 87.1 & 36.5 & 79.7 & - & - & - & 13.5 & 7.8 & 81.2 & 76.7 & 50.1 & 12.7 & 78.0 & 35.0 & 4.6 & 1.6 & - & 43.4 \\
PIT~\cite{PIT} &81.7 &26.9 &78.4 &6.3 &0.2 &19.8 &13.4 &17.4 &76.7 &74.1 &47.5 &22.4 &76.0 &21.7 &19.6 &27.7 &38.1 &44.9 \\ 
BDL~\cite{BDL} &72.0  &30.3  &74.5  &0.1 & 0.3  &24.6  &10.2  &\textbf{25.2}  &80.5  &80.0  &\textbf{54.7}  &\textbf{23.2}  &72.7  &24.0  &7.5  &\textbf{44.9} &39.0 &-\\

TGCF-DA + SE~\cite{choi2019self} & 90.1& 48.6& 80.7& 2.2& 0.2& 27.2& 3.2& 14.3& \textbf{82.1} & 78.4& 54.4& 16.4& 82.5& 12.3& 1.7& 21.8& 38.5& 46.6  \\
\midrule
\textbf{Ours}  & \textbf{93.1}& \textbf{53.2}& \textbf{81.1}& 2.6& \textbf{0.6}& \textbf{29.1}& 7.8& 15.7& 81.7 & 81.6& 53.6 & 20.1 & \textbf{82.7}& 22.9& 7.7& 31.3& \textbf{41.5} & \textbf{48.6}  \\
\bottomrule
\end{tabular}}
\end{table*}

\begin{table*}{}
\caption{Comparison results (mIoU) from GTAV to Cityscapes (with ResNet 101 backbone). }
\label{table:gtav_res}
\centering
\resizebox{\textwidth}{!}{%
\begin{tabular}{c|c|ccccccccccccccccccc|c}
\toprule
Method& Venue                 & \begin{turn}{90}road\end{turn} & \begin{turn}{90}sidewalk\end{turn} & \begin{turn}{90}building\end{turn} & \begin{turn}{90}wall\end{turn} & \begin{turn}{90}fence\end{turn} & \begin{turn}{90}pole\end{turn} & \begin{turn}{90}light\end{turn} & \begin{turn}{90}sign\end{turn} & \begin{turn}{90}vegetation\end{turn} & \begin{turn}{90}terrain\end{turn} & \begin{turn}{90}sky\end{turn} & \begin{turn}{90}person\end{turn} & \begin{turn}{90}rider\end{turn} & \begin{turn}{90}car\end{turn} & \begin{turn}{90}truck\end{turn} & \begin{turn}{90}bus\end{turn} & \begin{turn}{90}train\end{turn} & \begin{turn}{90}motocycle\end{turn} & \begin{turn}{90}bike\end{turn} & \begin{turn}{90}\textbf{mIoU}\end{turn}  \\ 
\toprule
Source Only &- &63.3 &15.7 &59.4 &8.6 &15.2 &18.3 &26.9 &15.0 &80.5 &15.3 &73.0 &51.0 &17.7 &59.7 &28.2 &33.1 &3.5 &23.2 &16.7 &32.9  \\ 
\midrule
BDL~\cite{BDL}&CVPR'19 &91.0 &44.7 &84.2 &34.6 &27.6 &30.2 &36.0 &36.0 &85.0 &43.6 &83.0 &58.6 &31.6 &83.3 &35.3 &49.7 &3.3 &28.8 &35.6 &48.5 \\
APODA~\cite{APODA}&AAAI'20  &85.6 &32.8 &79.0 &29.5 &25.5 &26.8 &34.6 &19.9 &83.7 &40.6 &77.9 &59.2 &28.3 &84.6 &34.6 &49.2 &8.0 &32.6 &39.6 &45.9\\
STAR~\cite{STAR}&CVPR'20 &88.4  &27.9  &80.8  &27.3  &25.6  &26.9  &31.6  &20.8  &83.5  &34.1  &76.6  &60.5  &27.2  &84.2  &32.9  &38.2  &1.0  &30.2  &31.2  &43.6\\
IntraDA~\cite{IntraDA}&CVPR'20 &90.6 &37.1 &82.6 &30.1 &19.1 &29.5 &32.4 &20.6 &85.7 &40.5 &79.7 &58.7 &31.1 &86.3 &31.5 &48.3 &0.0 &30.2 &35.8 &46.3 \\
SIM~\cite{SIM}&CVPR'20 &90.6 &44.7 &84.8 &34.3 &28.7 &31.6 &35.0 &37.6 &84.7 &43.3 &85.3 &57.0 &31.5 &83.8 &42.6 &48.5 &1.9 &30.4 &39.0 & 49.2 \\
LSE~\cite{LSE}&ECCV'20 &90.2 &40.0 &83.5 &31.9 &26.4 &32.6 &38.7 &37.5 &81.0 &34.2 &84.6 &61.6 &33.4 &82.5 &32.8 &45.9 &6.7 &29.1 &30.6 &47.5\\
WLabel~\cite{WLabel}&ECCV'20 &91.6 &47.4 &84.0 &30.4 &28.3 &31.4 &37.4 &35.4 &83.9 &38.3 &83.9 &61.2 &28.2 &83.7 &28.8 &41.3 &8.8 & 24.7 &46.4 &48.2 \\
CrCDA~\cite{CrCDA}&ECCV'20 &92.4 &55.3 &82.3 &31.2 &29.1 &32.5 &33.2 &35.6 &83.5 &34.8 &84.2 &58.9 &32.2 &84.7 &40.6 &46.1 &2.1 &31.1 &32.7 &48.6 \\
FADA~\cite{FADA}&ECCV'20 &92.5 &47.5 &85.1 &37.6 &32.8 &33.4 &33.8 &18.4 &85.3 &37.7 &83.5 &63.2 &39.7 &87.5 &32.9 &47.8 &1.6 &34.9 &39.5 & 49.2 \\
LDR~\cite{LDR}&ECCV'20 &90.8 &41.4 &84.7 &35.1 &27.5 &31.2 &38.0 &32.8 &85.6 &42.1 &84.9 &59.6 &34.4 &85.0 &42.8 &52.7 &3.4 &30.9 &38.1 &49.5\\
CCM~\cite{CCM}&ECCV'20 &93.5 &57.6 &84.6 &39.3 &24.1 &25.2 &35.0 &17.3 &85.0 &40.6 &86.5 &58.7 &28.7 &85.8 &49.0 &56.4 &5.4 &31.9 &43.2 &49.9 \\
CD-SAM~\cite{yang2021context} &WACV'21 &91.3 &46.0 &84.5 &34.4 &29.7 &32.6 &35.8 &36.4 &84.5 &43.2 &83.0 &60.0 &32.2 &83.2 &35.0 &46.7 &0.0 &33.7 &42.2 &49.2 \\
ASA~\cite{ASA} & TIP'21 &89.2 &27.8 &81.3 &25.3 &22.7 &28.7 &36.5 &19.6 &83.8 &31.4 &77.1 &59.2 &29.8 &84.3 &33.2 &45.6 &16.9 &34.5 &30.8 &45.1 \\
CLAN~\cite{CLANv2} & TPAMI'21 &88.7 &35.5 &80.3 &27.5 &25.0 &29.3 &36.4 &28.1 &84.5 &37.0 &76.6 &58.4 &29.7 &81.2 &38.8 &40.9 &5.6 &32.9 &28.8 &45.5\\
DAST~\cite{DAST}&AAAI'21 &92.2 &49.0 &84.3 &36.5 &28.9 &33.9 &38.8 &28.4 &84.9 &41.6 &83.2 &60.0 &28.7 &87.2 &45.0 &45.3 &7.4 &33.8 &32.8 &49.6 \\

\midrule

Ours& - &91.3 &48.6 &\textbf{85.5} &35.8 &31.4 &36.7 &37.5 &36.8 &\textbf{86.3} &40.3 &85.7 &\textbf{64.3} &31.1 &\textbf{87.7} &36.7 &44.9 &\textbf{15.9} &\textbf{38.9} &\textbf{55.4} &\textbf{51.9}\\

\bottomrule
 \end{tabular}}
\end{table*}

\begin{table}[h]{}
\caption{Segmentation results of Virtual KITTI $\rightarrow$ KITTI.}
\label{table:v2k_seg}
\centering
\begin{tabular}{l|c}
\hline
Method                &  \textbf{mIoU}\\ 
\toprule
GIO-Ada (CVPR'19)~\cite{GIO_Ada}  	&53.50 \\
\midrule
Self-Ensembling (ICCV'19)~\cite{choi2019self}  	&55.45  \\
\midrule
CutMix (BMVC'20)~\cite{french2019semi}   & 55.58  \\
\midrule
CowMix (Arxiv'20)~\cite{french2020milking}   & 56.07  \\
\midrule
DACS (WACV'21)~\cite{tranheden2020dacs}   & 55.51  \\
\midrule
PIT + CutMix (ICCV'21)~\cite{PIT} & 56.72 \\
PIT + CowMix (ICCV'21)~\cite{PIT} & 57.24 \\
PIT + DACS (ICCV'21)~\cite{PIT} & 56.57 \\
\midrule
Ours & \textbf{60.16} \\

\bottomrule

\end{tabular}
\end{table}

\begin{table}[h]{}
\caption{Segmentation results of Cityscapes $\rightarrow$ KITTI.}
\label{table:c2k_seg}
\centering

\begin{tabular}{l|c}
\hline
Method                 &  \textbf{mIoU}\\ 
\toprule

Self-Ensembling (ICCV'19) \cite{choi2019self}  &59.54  \\
\midrule
CutMix (BMVC'20) \cite{french2019semi}   & 58.78  \\
\midrule
CowMix (Arxiv'20) \cite{french2020milking}   & 59.15  \\
\midrule
DACS (WACV'21) \cite{tranheden2020dacs}   & 59.19  \\
\midrule
PIT + CutMix (ICCV'21)~\cite{PIT} & 60.09 \\
PIT + CowMix (ICCV'21)~\cite{PIT} & 60.37 \\
PIT + DACS (ICCV'21)~\cite{PIT} & 60.82 \\
\midrule
Ours & \textbf{61.62}\\

\bottomrule
\end{tabular}
\end{table}

\begin{table}[t]
\caption{Ablation of each component on SYNTHIA $\rightarrow$ Cityscapes.}
\label{table:ablation_component}
\centering
\begin{tabular}{ccc|c|c} \toprule
baseline &$M_{uncertainty}$  &$M_{classout}$  & mIoU & Gain\\
\midrule
$\surd$  &  &  & 51.5 & -\\
$\surd$  & $\surd$  &  & 53.5 & 2.0\\
$\surd$  & $\surd$  & $\surd$  & 55.9 & 4.4\\
\bottomrule
\end{tabular}
\end{table}

\begin{table}[t]
\caption{ Comparisons with the related work ~\cite{yu2019uncertainty}. on GTAV $\longrightarrow$ Cityscapes with ResNet 101 backbone.}
\label{table:comparison_60}
\begin{center}
\begin{tabular}{l|c|c} \toprule
method & \thead{mIoU \\ (GTAV)} & \thead{mIoU$_{13}$ \\ (SYN)}\\
\midrule
Mean Teacher~\cite{choi2019self}& 43.1 & 45.9\\
\midrule
+ Yu et al.~\cite{yu2019uncertainty} & 44.6 & 47.6\\
+ Ours  & \textbf{51.9} & \textbf{55.9}\\
\bottomrule
\end{tabular}
\end{center}
\end{table}

\setlength{\tabcolsep}{4pt}
\begin{table}
\begin{center}
\caption{Ablation study of each module's improvement from GTA5 to Cityscapes with VGG16 backbone. $L_{seg}$: Segmentation loss, $L_{mse}$: Mean Square Error used in \cite{choi2019self}, $L_{con}$: Our Uncertainty-Guided Consistency Loss,
$IT$: Image-to-Image translation for Style transfer.
}
\label{table:comparsion_choi}
\begin{tabular}{lllc}
\hline\noalign{\smallskip}
Method & Component & mIoU & Gain\\
\noalign{\smallskip}
\hline
\noalign{\smallskip}
Source Only & $L_{seg}$ & 28.3 & - \\
\hline
Choi \emph{et al.}~\cite{choi2019self} & $L_{seg}$+$L_{mse}$ & 32.6 & +4.3 \\
Ours (w/o $M_{classout}$)& $L_{seg}$+$L_{con}$ & 35.6 & +7.3 \\
\hline
Choi \emph{et al.}~\cite{choi2019self} & $L_{seg}$+$IT_1$~\cite{choi2019self} & 35.4 &+ 4.1\\
Ours (w/o $M_{classout}$)& $L_{seg}$+$IT_2$~\cite{BDL} & 35.1 & + 3.8 \\
\hline
Choi \emph{et al.}~\cite{choi2019self} & $L_{seg}$+$L_{mse}$+$IT_1$ & 42.5 &+14.2\\
Ours (w/o $M_{classout}$) & $L_{seg}$+$L_{con}$+$IT_2$ & 47.8 & +19.5 \\
\hline
\end{tabular}
\end{center}
\end{table}
\setlength{\tabcolsep}{1.4pt}

\begin{table*}[ht]
\caption{Comparison results (mIoU) from SYNTHIA to Cityscapes (with ResNet 101 backbone).}
\label{table:syn_res}
\centering
\begin{tabular}{c|c|ccccccccccccc|c} 
\toprule
Method & Venue               &  \begin{turn}{90}road\end{turn} & \begin{turn}{90}sidewalk\end{turn} & \begin{turn}{90}building\end{turn} & \begin{turn}{90}light\end{turn} & \begin{turn}{90}sign\end{turn} & \begin{turn}{90}vegetation\end{turn} & \begin{turn}{90}sky\end{turn} & \begin{turn}{90}person\end{turn} & \begin{turn}{90}rider\end{turn} & \begin{turn}{90}car\end{turn} & \begin{turn}{90}bus\end{turn} & \begin{turn}{90}motocycle\end{turn} & \begin{turn}{90}bike\end{turn} & \begin{turn}{90}\textbf{mIoU$_{13}$}\end{turn}  \\ 
\toprule
Source Only &-& 36.3 & 14.6 & 68.8 &5.6 &9.1 &69.0 &79.4 &52.5 &11.3 &49.8 &9.5 &11.0 &20.7 &29.5\\
\midrule
BDL~\cite{BDL}&CVPR'19 &86.0 &46.7 &80.3 &14.1 &11.6 &79.2 &81.3 &54.1 &27.9 &73.7 &42.2 &25.7 &45.3  &51.4 \\
DADA~\cite{DADA}&ICCV'19 &89.2 &44.8 &81.4 &8.6 &11.1 &81.8 &84.0 &54.7 &19.3 &79.7 &40.7 &14.0 &38.8  &49.8 \\
STAR~\cite{STAR} &CVPR'20 &82.6 &36.2 &81.1  &12.2 &8.7 &78.4 &82.2 &59.0 &22.5 &76.3 &33.6 &11.9 &40.8  &48.1 \\
IntraDA~\cite{IntraDA}&CVPR'20 &84.3 &37.7 &79.5 &9.2 &8.4 &80.0 &84.1 &57.2 &23.0 &78.0 &38.1 &20.3 &36.5 &48.9 \\
LTIR~\cite{LTIR}&CVPR'20 &92.6 &53.2 &79.2 &1.6 &7.5 &78.6 &84.4 &52.6 &20.0 &82.1 &34.8 &14.6 &39.4  &49.3 \\
SIM~\cite{SIM}&CVPR'20 &83.0 &44.0 &80.3 & 17.1 &15.8 &80.5 &81.8 &59.9 &33.1 &70.2 &37.3 &28.5 &45.8  &52.1 \\
LSE~\cite{LSE}&ECCV'20 &82.9 &43.1 &78.1 &9.1 &14.4 &77.0 &83.5 &58.1 &25.9 &71.9 &38.0 &29.4 &31.2  &49.4 \\
CrCDA~\cite{CrCDA}&ECCV'20 &86.2 &44.9 &79.5 &9.4 &11.8 &78.6 &86.5 &57.2 &26.1 &76.8 &39.9 &21.5 &32.1  &50.0 \\
WLabel~\cite{WLabel}&ECCV'20 &92.0 &53.5 &80.9 &3.8 &6.0 &81.6 &84.4 &60.8 &24.4 &80.5 &39.0 &26.0 &41.7  &51.9 \\

CD-SAM~\cite{yang2021context} &WACV'21 &82.5 &42.2 &81.3 &18.3 &15.9 &80.6 &83.5 &61.4 &33.2 &72.9 &39.3 &26.6 &43.9 &52.4\\

CLAN~\cite{CLANv2} &TPAMI'21 &82.7 &37.2 &81.5 &17.1 &13.1 &81.2 &83.3 &55.5 &22.1 &76.6 &30.1 &23.5 &30.7 &48.8\\

ASA~\cite{ASA} &TIP'21 &91.2 &48.5 &80.4 &5.5 &5.2 &79.5 &83.6 &56.4 &21.9 &80.3 &36.2 &20.0 &32.9 &49.3\\

DAST~\cite{DAST}&AAAI'21 &87.1 &44.5 &82.3 &13.9 &13.1
&81.6 &86.0 &60.3 &25.1 &83.1 &40.1 &24.4 &40.5 &52.5\\
\midrule

Ours &- &85.5 &42.5 &\textbf{83.0} &\textbf{20.9} &\textbf{25.5} &\textbf{82.5} &\textbf{88.0} &\textbf{63.2} &\textbf{31.8} &\textbf{86.5} &\textbf{41.2} &25.9 &\textbf{50.7}  &\textbf{55.9}\\

\bottomrule
\end{tabular}
\end{table*}

\subsection{Datasets}
\noindent \textbf{Cityscapes}~\cite{cordts2016cityscapes} is a dataset focused on autonomous driving,
which consists of 2,975 images in the training
set, and 500 images in the validation set. The images have
a fixed spatial resolution of 2048 $\times$ 1024 pixels. 
For the sake of the fairness of experimental results, we follow the same evaluation protocol~\cite{AdaptSegNet,advent,CLAN}, i.e. we train the model on the unlabeled training set and report the results on the validation set.

\noindent \textbf{GTAV}~\cite{gtav} is a synthetic dataset including 24,966 photo-realistic images rendered by the gaming engine Grand Theft Auto V (GTAV).
The resolution of images is  1914 $\times$ 1051 pixels
which is similar to Cityscapes that the semantic categories
are also compatible between the two datasets. We use all the 19 official training classes in our experiments.

\noindent \textbf{SYNTHIA}~\cite{synthia} is another synthetic dataset composed of 9,400
annotated synthetic images with the resolution 1280 $\times$ 960.
Like GTAV, it has semantically compatible annotations with
Cityscapes. Following the prior works~\cite{CrossCity,CDA,ROAD},
we use the SYNTHIA-RAND-CITYSCAPES subset~\cite{synthia} as our training set.

\noindent \textbf{KITTI} \cite{kITTI}  is a real-world dataset containing 7,481 images with bounding boxes and another 200 images with pixel-level labels. In the detection task, we split the training set and the validation set manually with a ratio of $9:1$ following \cite{PIT}. 
In the segmentation task, it is used as the target domain only due to the lack of pixel-level annotations. 

\noindent \textbf{Virtual KITTI} \cite{VKITTI} is a synthetic dataset which clones the scenes from the KITTI with 21,260 images.  Each image is densely annotated at pixel level with category and depth information. 
It is designed to mimic the conditions of  KITTI dataset and 
has similar scene layouts, camera viewpoints and image resolution to KITTI dataset.

\subsection{Implementation details}
Following common UDA protocols~\cite{choi2019self,SEANet,AdaptSegNet,CLAN}, we employ the VGG-16~\cite{VGG} and ResNet 101~\cite{he2016deep} as the backbone of the DeepLab-v2~\cite{chen2018deeplab} in our implementations, and the backbone model is pre-trained on ImageNet \cite{imagenet}. For the DeepLab-v2 network, we use Adam as the optimizer. The initial learning rate is  $1 \times 10 ^{-5}$, and the weight decay is  $5 \times 10 ^{-5}$.  
In our uncertainty module, we perform $N=8$ times stochastic forward passes to capture the understanding of latent epistemic uncertainty. We set the EMA decay $\alpha$ to 0.999 during the training process. Following~\cite{laine2016temporal,Mean_teacher,choi2019self},the consistency weight is a ramp-up function: $\lambda_{con}=\lambda_0 \times  e^{-5(1-t/t_{max})^2} $, where $\lambda_0$ is an initial constant. This time-dependent threshold function is used to increase the certainty at later training steps. We set $\alpha=0.75$ and $\beta=-5$ in all experiments.
Our method is implemented in Pytorch on a single NVIDIA GTX 3090 Ti.

\subsection{Comparisons with the State-of-the-art Techniques}
We compare the results
between our method and the state-of-the-art methods on four challenging benchmarks, which includes the synthetic-to-real adaptation, \emph{i.e.,}
``GTAV $\rightarrow$ Cityscapes'' and ``SYNTHIA $\rightarrow$ Cityscapes'', ``Virtual KITTI $\rightarrow$ KITTI'' and cross-city adaptation, \emph{i.e.,} ``Cityscapes $\rightarrow$ KITTI''. With VGG16 backbone, our proposed method significantly outperforms the state-of-the-art methods by $5\% \sim 8\%$ on GTAV $\rightarrow$ Cityscapes,  and $2\% \sim 7\%$ on SYNTHIA $\rightarrow$ Cityscapes. Besides, it is superior to the non-adaptive baseline by $19.5\%$ on GTA5 $\rightarrow$ Cityscapes and $20\% \sim 24\%$ on SYNTHIA $\rightarrow$ Cityscapes. With ResNet101 backbone, our proposed method outperforms the state-of-the-art methods by $1\% \sim 3\%$ on GTAV $\rightarrow$ Cityscapes,  and $2\% \sim 6\%$ on SYNTHIA $\rightarrow$ Cityscapes.

\subsubsection{Results on GTAV $\rightarrow$ Cityscapes}
As shown in Table~\ref{table:gtav} and Table~\ref{table:gtav_res}, we present the adaptation results from GTAV to Cityscapes with VGG16 and ResNet 101, respectively. Source-only denotes the baseline Deeplab-v2~\cite{chen2018deeplab} is trained
with only source domain data.
In the works~\cite{CDA,AdaptSegNet,ROAD,CLAN,APODA,IntraDA,FADA}, they mainly focused on distribution alignment via different adversarial mechanisms. But promoting feature alignment only on the high representation level is not enough, \emph{i.e.,} feature level~\cite{CDA,ROAD} or output level~\cite{AdaptSegNet,CLAN,SIM}. The best results of mIoU among them are still about $7\%$ worse than our results. To further reduce the domain gap, Hoffman \emph{et. al} ~\cite{Cycada} introduced an image-to-image translation model to perform a style transfer process on the low appearance level. Such techniques are further integrated into \cite{SIM,BDL,LSD,Conservative_loss,choi2019self} to achieve higher performance, while they are still about $5\% \sim 10\%$ worse than our results.
Another line of non-adversarial methods~\cite{CDA,CBST,advent} were proposed to address the negative effect of adversarial training. The self-ensembling with GAN-based augmentation~\cite{choi2019self} has been recently proposed and surpassed most of the previous works. In Table~\ref{table:gtav}, our method could get about $5.3\%$ improvements compared to this work~\cite{choi2019self}. Extensive experiments in Table~\ref{table:gtav} and Table~\ref{table:gtav_res} show that our approach achieves a new top performance.

\subsubsection{Results on SYNTHIA $\rightarrow$ Cityscapes}
As shown in Table~\ref{table:synthia} and Table~\ref{table:syn_res}, we list the adaptation results on the task ''SYNTHIA $\rightarrow$ Cityscapes'' with VGG16 and ResNet 101, respectively.  Due to the fact that the baselines~\cite{AdaptSegNet,CLAN} only calculate the results using 13 categories, we also list results for the 13 categories for a fair comparison. Although the domain gap between SYNTHIA and Cityscapes is much larger than that of GTAV to Cityscapes, we could observe in Table~\ref{table:synthia} that our uncertainty-aware consistency regularization still performs well in terms of both mIoU and per-class IoU. In some semantic categories, such as large objects, \emph{e.g.,} road, building, wall, vegetation, sky, \emph{etc.}, our method could capture the understanding of epidemic uncertainty and remarkably increase the certainty of these categories during the training procedure. In Table~\ref{table:synthia}, the proposed method significantly outperforms the state-of-the-art techniques by $2.5\%$ in mIoU16 and $2\%$ in mIoU13 with VGG16 backbone. It is superior to the non-adaptive baseline by $18.9\%$ in mIoU16 and $24.5\%$ in mIoU13. In Table~\ref{table:syn_res}, the proposed method outperforms the state-of-the-art approaches by $2\% \sim 5\%$ with ResNet101 backbone. 

\subsubsection{Virtual KITTI $\rightarrow$ KITTI and Cityscapes $\rightarrow$ KITTI} In addition to the two commonly-used benchmarks, we also conduct experiments on another synthetic-to-real adaptation, \emph{i.e.,}  Virtual KITTI $\rightarrow$ KITTI, and cross-city adaptation, \emph{i.e.,} Cityscapes $\rightarrow$ KITTI, to validate the effectiveness of our method. Table \ref{table:v2k_seg} shows the results of adapting the model from Virtual KITTI to KITTI. We reproduce Self-Ensembling~\cite{choi2019self}, CutMix~\cite{french2019semi}, CowMix~\cite{french2020milking}, DACS~\cite{tranheden2020dacs} in the same setting. The results of GIO-Ada~\cite{GIO_Ada} and PIT~\cite{PIT} are reported in the original papers. We can see that our method significantly improves the mIoU by $3.4\% \sim 6.6\%$ compared with the existing UDA methods. In Table~\ref{table:c2k_seg}, we adapt from Cityscapes to KITTI, where the source domains and target domain have different distributions in cross-city road scenes and street views. Our proposed method can outperform the state-of-the-art methods by around $1\%$. 
Table~\ref{table:c2k_seg} demonstrate our effectiveness in cross-city adaptation.

\subsection{Ablation Study}
\noindent \textbf{Ablation of each component:} In Table~\ref{table:ablation_component}, we investigate the effects of different design components in SYNTHIA $\rightarrow$ Cityscapes with ResNet101 backbone. The uncertainty mask $M_{uncertainty}$ and the ClassDrop mask $M_{classout}$ reveals the contribution of the proposed dynamic weighting scheme and the ClassOut strategy are complementary. The consistency regularization baseline is $51.5\%$. 
By adding the $M_{uncertainty}$ and $M_{classout}$ sequentially, we boost the mIoU with an additional $+2.0\%$ and $+2.4\%$, achieving $53.5\%$ and $55.9\%$, respectively. These improvements show the effects of individual components of our proposed approach. 

\begin{figure*} 
\centering
\includegraphics[scale=1.3]{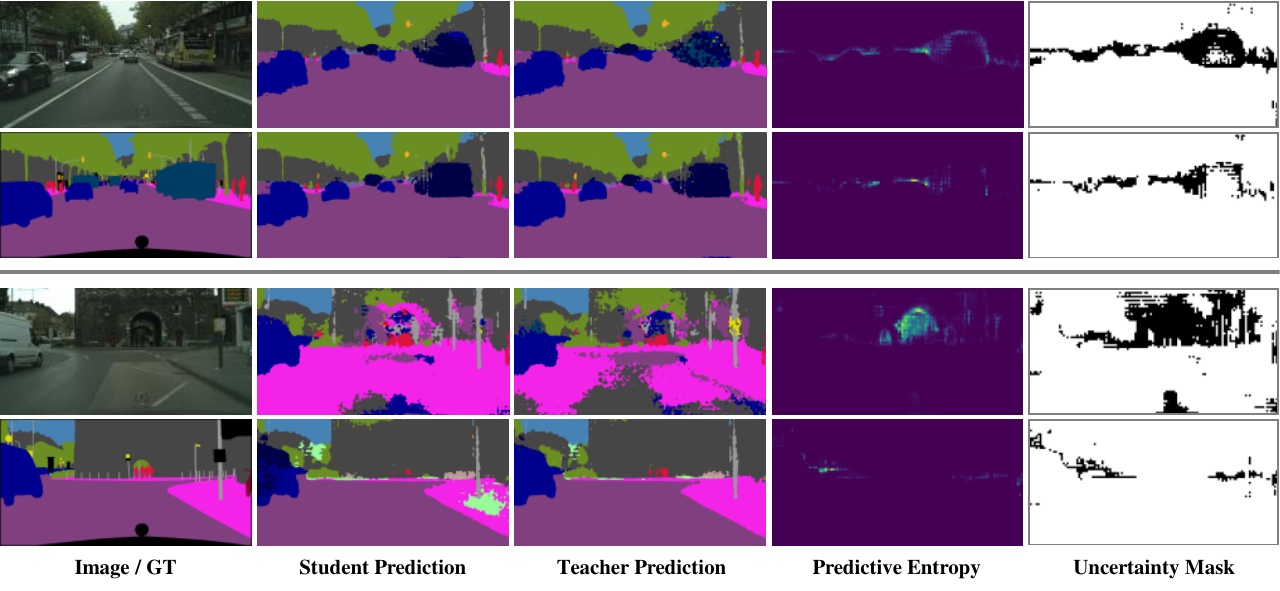}
\caption{ Visualization results of GTA5 $\rightarrow$ Cityscapes (first and second rows) and SYNTHIA $\rightarrow$ Cityscapes (third and fourth rows).
Segmentation results at 10K training steps (first and third rows) and 56K training steps (second and fourth rows). The fourth and fifth
columns illustrate the predictive entropy and our uncertainty mask.}
\label{fig: Visualization}
\end{figure*}

 \noindent \textbf{Comparison to the related work~\cite{yu2019uncertainty} :} In Table. \ref{table:comparison_60}, we show the experimental comparison on two benchmark datasets with ResNet 101 backbones to demonstrate its effectiveness.
Note that all the experimental results of Table. \ref{table:comparison_60} are conducted on the same Mean-Teacher baseline with ResNet 101 backbones. We replace the proposed method with the approach~\cite{yu2019uncertainty}, and we find that the improvements of~\cite{yu2019uncertainty} are limited over the Mean Teacher baseline, only achieving 44.6 and 47.6 in GTAV $\rightarrow$ Cityscapes and SYNTHIA $\rightarrow$ Cityscapes, respectively.
Our proposed method outperforms the related work~\cite{yu2019uncertainty} by 7.3 $\%$ and 8.3 $\%$ on GTAV $\rightarrow$ Cityscapes and SYNTHIA $\rightarrow$ Cityscapes, achieving $51.9\%$ and $55.9\%$, respectively. 

 \noindent \textbf{Comparison to the related work~\cite{choi2019self} :} In Table \ref{table:comparsion_choi}, we compare our method with the non-adaptive baseline and Self-Ensembling (SE)~\cite{choi2019self} with VGG16 backbone. $L_{seg}$ denotes the supervised segmentation loss, $L_{mse}$ refers to the common Mean Square Error used in~\cite{choi2019self}, and $L_{con}$ is our uncertainty-guided consistency Loss with the dynamic weighting scheme. As we can see, the Source Only baseline achieves $28.3\%$ from GTAV dataset to Cityscapes dataset. We see that in the third row, Choi \emph{et al.} achieves a performance of $32.6\%$ in the original consistency loss ($L_{seg}+L_{mse}$). Our uncertainty-guided consistency loss achieves about $3.0\%$ improvement over directly using the Mean Square Error ($L_{seg}+L_{con}$), reaching  $35.6\%$ in mIoU.

As mentioned in Section 2, pixel-level adaptation is also a
key factor in minimizing the discrepancy of data distribution. Therefore, it is helpful to utilize a transferred source domain image dataset whose appearance is more similar to that of the target-domain image dataset. Following common practice~\cite{SIM,BDL}, we adopt the transferred GTA5 images of \cite{BDL} which utilizes a CycleGAN\cite{CycleGAN2017} structure to adapt the style of GTAV images to the style of Cityscapes images. In the fifth row and sixth row of Table 3, we could find that our Image-to-Image Translation achieves a similar performance compared to \cite{choi2019self}.
On top of that, as we can see in the last row, our final adaptive performance is superior to the state-of-the-art by $5.3\%$, resulting in a $19.5\%$ increase in mIoU over the non-adaptive baseline. 

\begin{figure} 
\includegraphics[scale=1.8]{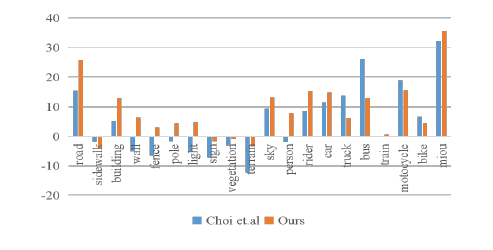}
\caption{ Comparisons of Per-Class IoU Gain between Choi et.al \cite{choi2019self} and ours w/o IT with VGG16 backbone in GTAV $\rightarrow$ Cityscapes.}
\label{fig:per_class_iou}
\end{figure}

\begin{figure} 
\includegraphics[scale=0.6]{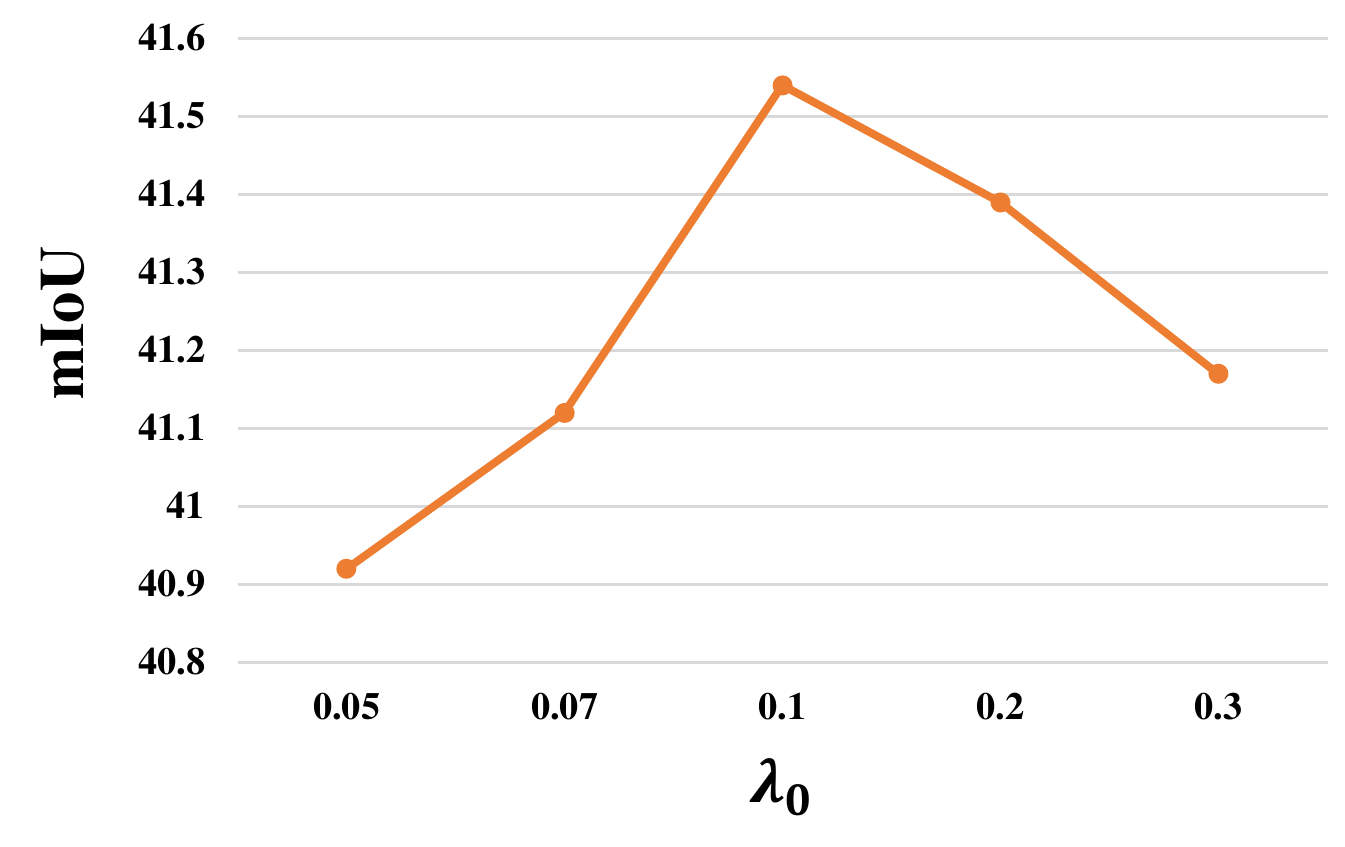}
\caption{Parameter analysis about $\lambda_0$ (SYNTHIA $\rightarrow$ Cityscapes). }
\label{fig:param}
\end{figure}

\subsection{Analysis}
In this section, we provide visualization results and provide some analysis of our proposed framework.

Fig.~\ref{fig:per_class_iou} shows the comparison results of the per-class IoU gain and comparisons of mIoU between the SE baseline~\cite{choi2019self} and our proposed method. In many large categories, \emph{i.e.,} road, building, sky, that have long boundaries, we have achieved a per-class IoU performance improvement. In other static categories, such as sidewalk, sign, vegetation and terrain, our method achieves a lower performance degradation than \cite{choi2019self}. Besides, as we can see in Table \ref{table:gtav_res} and Table \ref{table:syn_res}, our method shows good performance in some moving objects, \emph{e.g.,} motorcycle, bicycle, etc. 
Our method obtains an overall better performance than ~\cite{choi2019self}.


\begin{figure*} 
\centering
\includegraphics[scale=1.0]{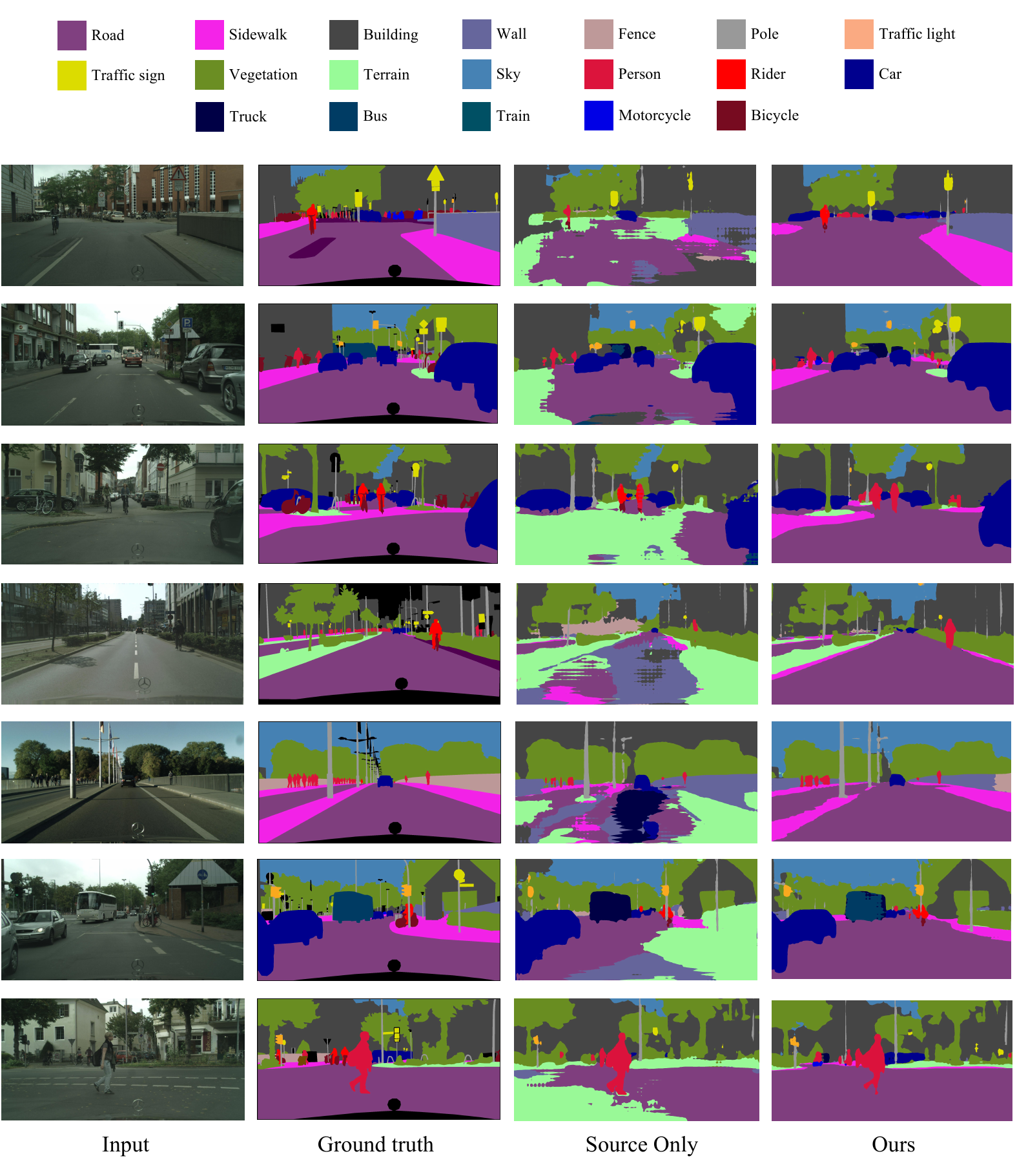}
\caption{ Semantic segmentation qualitative results from GTA5 to Cityscapes. From left to right: target image, ground truth, source-only prediction, and predictions using our method.}
\label{fig:results}
\end{figure*}

\subsubsection{Visualization}
The effectiveness of the uncertainty-aware consistency regularization  is shown in Fig.~\ref{fig: Visualization}. We visualize the student prediction, teacher prediction, the entropy map of the teacher model, and our uncertainty mask. In  Fig.~\ref{fig: Visualization}, as we can see in the fourth column, the predictive entropy captures the latent epidemic uncertainty, especially for some specific large objects, such as car and truck. In the fifth column, the white pixels of the uncertainty mask are the ones with higher confidence. The first and third rows show that our uncertainty mask effectively filters out the unreasonable pixels and guides the teacher to be a good proxy for training the student network in the early stage of the training process. In addition, the second and fourth rows show that our uncertainty module pays attention to the semantic boundary of objects in the later training stage. 

These qualitative results are consistent with our motivations and reveal the reason why current consistency regularization methods are often unstable in minimizing the distribution discrepancy, which lies in two aspects. Firstly, directly imposing a simple MSE constraint as consistency loss onto all pixels could harm the learning process by generating unreasonable guidance from the teacher to the student model. Secondly, due to the  ``error accumulation'' in the teacher model, it could take more training iterations to converge and even may lead to early performance degradation during the adaptation process. In the second row of Figure 4, the entropy is low and not obvious to human naked eyes.  However, our uncertainty mask can still filter out these uncertain areas. 
The main reason is that our dynamic weighting scheme can enable dynamic adjustion during the training. When the entropy is low, the dynamic threshold will be updated accordingly, and thus, we can filter out theses unobvious areas.

In Fig.~\ref{fig:results}, we illustrate some qualitative results of our models tested on the validation sets of Cityscapes~\cite{cordts2016cityscapes} dataset. Following prior works~\cite{SEANet,Cycada,CAG_UDA}, we show the target image, ground truth,
source only prediction, and our prediction from left to right. Without domain adaptation, the model trained only on source supervision produces noisy segmentation predictions. With the help of our uncertainty-aware consistency-regularization,  our method manages to produce correct predictions at a high level of confidence.

\subsubsection{Parameter Analysis}
In this section, we investigate the sensitivity of hyper-parameters $\lambda_0$, $\alpha$, $\beta$, $N$ and augmentations. 
 
\noindent \textbf{Effect of $\lambda_0$:} $\lambda_0$ means the initial state of consistency weight $\lambda_{con}$, which balances the domain adaptation process among different loss functions, and it is crucial in the training process. 
 In Fig.~\ref{fig:param}, the best performance from SYNTHIA to Cityscapes (w VGG16 backbone) occurs when the initial value of $\lambda_0$ is $0.1$. The results of mean mIoU over the 16 common classes are reported.

 \noindent \textbf{Effect of $\alpha$:} In this experiment, we set  $\beta=-5$, $N=8$ to check the sensitivity of $\alpha$ in SYNTHIA to Cityscapes (w ResNet101 backbone). $\alpha$ is the initial state of the dynamic threshold in Eq \ref{eq:thres}. In Table \ref{table:para_alpha}, we find that when $\alpha$ is set to 0.725, we can obtain the highest performance.

\noindent \textbf{Effect of  $\beta$:} In this experiment, we adapt from SYNTHIA to Cityscapes (w ResNet101 backbone) to discuss the selection of the parameter $\beta$, which controls the exponential speed of the dynamic threshold in Equation \ref{eq:thres}. We set the other parameters $ \alpha= 0.75$,  $N=8$. In Table \ref{table:para_beta}, we observe that the highest mIoU on target domain is achieved when the value of $\beta$ is around $-4.8$, which means that, under such condition, the exponential speed benefits the dynamic threshold of the domain adaptation the most.
  
\noindent \textbf{Effect of $N$:} In this part, we analyze the selection of $N$, which is the copy numbers of target image in the stochastic forward pass. We set the other parameters as: $ \alpha= 0.75$, $ \beta= -5$. In Table \ref{table:para_N}, we find that when $N$ is set to 8, it achieves the best performance in SYNTHIA to Cityscapes (w ResNet101 backbone). Therefore, $N$ is set to 8 in all experiments.
   
\setlength{\tabcolsep}{4pt}
\begin{table}
\begin{center}
\caption{Parameter analysis about $\alpha$. }

\label{table:para_alpha}
\begin{tabular}{|c|c|c|c|c|c|c|c|}
\hline
$\alpha$ & 0.675 & 0.70 & 0.725 & 0.75 & 0.775 & 0.80 \\
\hline
mean IoU &54.0  & 53.9 &\textbf{55.1} &54.5 & 54.7  & 53.3 \\
\hline
\end{tabular}

\end{center}
\end{table}
\setlength{\tabcolsep}{1.4pt}

\setlength{\tabcolsep}{4pt}
\begin{table}
\begin{center}
\caption{Parameter analysis about $\beta$. }

\label{table:para_beta}
\begin{tabular}{|c|c|c|c|c|c|c|c|c|}
\hline
$\beta$ & -5.3 & -5.2 & -5.1 & -5.0 & -4.9 & -4.8 & -4.7  \\
\hline
mean IoU & 54.2 & 54.7 & 53.7 &54.5 &54.7 & \textbf{55.9} & 54.5   \\
\hline
\end{tabular}
\end{center}
\end{table}
\setlength{\tabcolsep}{1.4pt}

\setlength{\tabcolsep}{4pt}
\begin{table}
\begin{center}
\caption{Parameter analysis about $N$. }

\label{table:para_N}
\begin{tabular}{|c|c|c|c|c|c|c|}
\hline
$N$  & 6 & 7 & 8 & 9 & 10   \\
\hline
mean IoU  & 53.6 & 54.0 &\textbf{54.5} &53.4 & 54.1    \\
\hline
\end{tabular}
\end{center}
\end{table}
\setlength{\tabcolsep}{1.4pt}

 \noindent \textbf{Effects of Augmentations:} We investigate the sensitivity of augmentation, \emph{e.g,} Gaussian noise, color jittering, random crop. GN, CJ, RC are the abbreviation of Gaussian noise, color jittering, random crop, respectively. The ablations of each augmentation are shown in Table \ref{table:ablation_augmentation} when adapting from  SYNTHIA to Cityscapes (w ResNet101 backbone). From the Table \ref{table:ablation_augmentation}, we can observe that  augmentations are complementary and our consistency regularization needs to be conducted under the condition that the student image and the target image are imposed with different augmentations. 
 
\begin{table}[t]
\caption{Ablation study of each augmentation. }
\label{table:ablation_augmentation}
\begin{center}
\begin{tabular}{cccc|c} \toprule
No Aug & RC & GN & CJ  & mIoU \\
\midrule
$\surd$  &  &   & & 52.2 \\
$\surd$  & $\surd$ & &  & 51.9\\
$\surd$  & $\surd$ & $\surd$ & & 53.7\\
$\surd$  & $\surd$ & $\surd$ & $\surd$ & 55.9 \\
\bottomrule
\end{tabular}
\end{center}
\end{table}
 
 \subsection{Limitations}
  \noindent 1) We develop a unified uncertainty-aware consistency regularization in this work. Though our method has achieved very good results, it can hardly treat the stuff regions and the instances of things in a different manner to reduce the uncertainty. 
 2) The scale-invariant feature across different frames are neglected in this work, which can be utilized as prior knowledge for effective domain adaptation for video semantic segmentation.  
 As future work, these interesting points will be investigated.
 
\section{Conclusion} 

In this paper, we proposed an uncertainty-aware consistency regularization technique to address the domain shift for cross-domain segmentation. Our uncertainty module is capable of estimating the latent uncertainty map for the purpose of a better knowledge transfer. Specifically, We first introduced an uncertainty-guided consistency loss with a dynamic weighting scheme for filtering out the unreasonable pixels and mining the high confident predictions of target samples. Secondly, we present a ClassDrop mask generation algorithm to generate class-wise perturbations. Guided by this mask, we present a ClassOut strategy to keep the local regional consistency in varying environments.
Experimental results verify that our method is superior to existing state-of-the-art approaches on four challenging benchmark datasets. 

\section{Acknowledgments}
This work is supported by National Key Research and Development Program of China (No. 2019YFC1521104), National Natural Science Foundation of China (No. 61972157). The author Qianyu Zhou is supported by Wu Wenjun Honorary Doctoral Scholarship, AI Institute, Shanghai Jiao Tong University. Also, the authors would like to thank Chuyun Zhuang (Shanghai Jiao Tong University) for his efforts, and Zhe Hu (Hikvision) and Xin Tan (Shanghai Jiao Tong University) for their valuable suggestions.

\bibliographystyle{model2-names_clean}
\bibliography{refs_clean}

\begin{thebibliography}{89}
\expandafter\ifx\csname natexlab\endcsname\relax\def\natexlab#1{#1}\fi
\providecommand{\url}[1]{\texttt{#1}}
\providecommand{\href}[2]{#2}
\providecommand{\path}[1]{#1}
\providecommand{\DOIprefix}{doi:}
\providecommand{\ArXivprefix}{arXiv:}
\providecommand{\URLprefix}{URL: }
\providecommand{\Pubmedprefix}{pmid:}
\providecommand{\doi}[1]{\href{http://dx.doi.org/#1}{\path{#1}}}
\providecommand{\Pubmed}[1]{\href{pmid:#1}{\path{#1}}}
\providecommand{\bibinfo}[2]{#2}
\ifx\xfnm\relax \def\xfnm[#1]{\unskip,\space#1}\fi
\bibitem[{Cariucci et~al.(2017)Cariucci, Porzi, Caputo, Ricci and
  Bulo}]{cariucci2017autodial}
\bibinfo{author}{Cariucci, F.M.}, \bibinfo{author}{Porzi, L.},
  \bibinfo{author}{Caputo, B.}, \bibinfo{author}{Ricci, E.},
  \bibinfo{author}{Bulo, S.R.}, \bibinfo{year}{2017}.
\newblock \bibinfo{title}{Autodial: Automatic domain alignment layers}, in:
  \bibinfo{booktitle}{ICCV}.
\bibitem[{Chang et~al.(2019)Chang, Wang, Peng and Chiu}]{DISE}
\bibinfo{author}{Chang, W.L.}, \bibinfo{author}{Wang, H.P.},
  \bibinfo{author}{Peng, W.H.}, \bibinfo{author}{Chiu, W.C.},
  \bibinfo{year}{2019}.
\newblock \bibinfo{title}{All about structure: Adapting structural information
  across domains for boosting semantic segmentation}, in:
  \bibinfo{booktitle}{Proceedings of the IEEE/CVF Conference on Computer Vision
  and Pattern Recognition}, pp. \bibinfo{pages}{1900--1909}.
\bibitem[{Chen et~al.(2018a)Chen, Papandreou, Kokkinos, Murphy and
  Yuille}]{chen2018deeplab}
\bibinfo{author}{Chen, L.}, \bibinfo{author}{Papandreou, G.},
  \bibinfo{author}{Kokkinos, I.}, \bibinfo{author}{Murphy, K.},
  \bibinfo{author}{Yuille, A.L.}, \bibinfo{year}{2018}a.
\newblock \bibinfo{title}{Deeplab: Semantic image segmentation with deep
  convolutional nets, atrous convolution, and fully connected crfs}.
\newblock \bibinfo{journal}{{IEEE} Trans. Pattern Anal. Mach. Intell.}
  \bibinfo{volume}{40}, \bibinfo{pages}{834--848}.
\bibitem[{Chen et~al.(2018b)Chen, Zhu, Papandreou, Schroff and
  Adam}]{chen2018encoder}
\bibinfo{author}{Chen, L.C.}, \bibinfo{author}{Zhu, Y.},
  \bibinfo{author}{Papandreou, G.}, \bibinfo{author}{Schroff, F.},
  \bibinfo{author}{Adam, H.}, \bibinfo{year}{2018}b.
\newblock \bibinfo{title}{Encoder-decoder with atrous separable convolution for
  semantic image segmentation}, in: \bibinfo{booktitle}{Proceedings of the
  European conference on computer vision (ECCV)}, pp.
  \bibinfo{pages}{801--818}.
\bibitem[{Chen et~al.(2019a)Chen, Li, Chen and Gool}]{GIO_Ada}
\bibinfo{author}{Chen, Y.}, \bibinfo{author}{Li, W.}, \bibinfo{author}{Chen,
  X.}, \bibinfo{author}{Gool, L.V.}, \bibinfo{year}{2019}a.
\newblock \bibinfo{title}{Learning semantic segmentation from synthetic data: A
  geometrically guided input-output adaptation approach}, in:
  \bibinfo{booktitle}{Proceedings of the IEEE/CVF Conference on Computer Vision
  and Pattern Recognition}, pp. \bibinfo{pages}{1841--1850}.
\bibitem[{Chen et~al.(2018c)Chen, Li and Van~Gool}]{ROAD}
\bibinfo{author}{Chen, Y.}, \bibinfo{author}{Li, W.},
  \bibinfo{author}{Van~Gool, L.}, \bibinfo{year}{2018}c.
\newblock \bibinfo{title}{Road: Reality oriented adaptation for semantic
  segmentation of urban scenes}, in: \bibinfo{booktitle}{Proceedings of the
  IEEE conference on computer vision and pattern recognition}, pp.
  \bibinfo{pages}{7892--7901}.
\bibitem[{Chen et~al.(2019b)Chen, Lin, Yang and Huang}]{chen2019crdoco}
\bibinfo{author}{Chen, Y.C.}, \bibinfo{author}{Lin, Y.Y.},
  \bibinfo{author}{Yang, M.H.}, \bibinfo{author}{Huang, J.B.},
  \bibinfo{year}{2019}b.
\newblock \bibinfo{title}{Crdoco: Pixel-level domain transfer with cross-domain
  consistency}, in: \bibinfo{booktitle}{CVPR}.
\bibitem[{Chen et~al.(2017)Chen, Chen, Chen, Tsai, Frank~Wang and
  Sun}]{CrossCity}
\bibinfo{author}{Chen, Y.H.}, \bibinfo{author}{Chen, W.Y.},
  \bibinfo{author}{Chen, Y.T.}, \bibinfo{author}{Tsai, B.C.},
  \bibinfo{author}{Frank~Wang, Y.C.}, \bibinfo{author}{Sun, M.},
  \bibinfo{year}{2017}.
\newblock \bibinfo{title}{No more discrimination: Cross city adaptation of road
  scene segmenters}, in: \bibinfo{booktitle}{Proceedings of the IEEE
  International Conference on Computer Vision}, pp.
  \bibinfo{pages}{1992--2001}.
\bibitem[{Choi et~al.(2019)Choi, Kim and Kim}]{choi2019self}
\bibinfo{author}{Choi, J.}, \bibinfo{author}{Kim, T.}, \bibinfo{author}{Kim,
  C.}, \bibinfo{year}{2019}.
\newblock \bibinfo{title}{Self-ensembling with gan-based data augmentation for
  domain adaptation in semantic segmentation}, in:
  \bibinfo{booktitle}{Proceedings of the IEEE/CVF International Conference on
  Computer Vision}, pp. \bibinfo{pages}{6830--6840}.
\bibitem[{Cordts et~al.(2016)Cordts, Omran, Ramos, Rehfeld, Enzweiler,
  Benenson, Franke, Roth and Schiele}]{cordts2016cityscapes}
\bibinfo{author}{Cordts, M.}, \bibinfo{author}{Omran, M.},
  \bibinfo{author}{Ramos, S.}, \bibinfo{author}{Rehfeld, T.},
  \bibinfo{author}{Enzweiler, M.}, \bibinfo{author}{Benenson, R.},
  \bibinfo{author}{Franke, U.}, \bibinfo{author}{Roth, S.},
  \bibinfo{author}{Schiele, B.}, \bibinfo{year}{2016}.
\newblock \bibinfo{title}{The cityscapes dataset for semantic urban scene
  understanding}, in: \bibinfo{booktitle}{Proc. CVPR}, pp.
  \bibinfo{pages}{3213--3223}.
\bibitem[{Csurka(2017)}]{csurka2017domain}
\bibinfo{author}{Csurka, G.}, \bibinfo{year}{2017}.
\newblock \bibinfo{title}{Domain adaptation for visual applications: A
  comprehensive survey}.
\newblock \bibinfo{journal}{arXiv preprint arXiv:1702.05374} .
\bibitem[{Deng et~al.(2009)Deng, Dong, Socher, Li, Li and Fei-Fei}]{imagenet}
\bibinfo{author}{Deng, J.}, \bibinfo{author}{Dong, W.},
  \bibinfo{author}{Socher, R.}, \bibinfo{author}{Li, L.J.},
  \bibinfo{author}{Li, K.}, \bibinfo{author}{Fei-Fei, L.},
  \bibinfo{year}{2009}.
\newblock \bibinfo{title}{Imagenet: A large-scale hierarchical image database},
  in: \bibinfo{booktitle}{CVPR}.
\bibitem[{Du et~al.(2019)Du, Tan, Yang, Feng, Xue, Zheng, Ye and
  Zhang}]{du2019ssf}
\bibinfo{author}{Du, L.}, \bibinfo{author}{Tan, J.}, \bibinfo{author}{Yang,
  H.}, \bibinfo{author}{Feng, J.}, \bibinfo{author}{Xue, X.},
  \bibinfo{author}{Zheng, Q.}, \bibinfo{author}{Ye, X.},
  \bibinfo{author}{Zhang, X.}, \bibinfo{year}{2019}.
\newblock \bibinfo{title}{Ssf-dan: Separated semantic feature based domain
  adaptation network for semantic segmentation}, in: \bibinfo{booktitle}{ICCV}.
\bibitem[{Everingham et~al.(2010)Everingham, Van~Gool, Williams, Winn and
  Zisserman}]{Pascal}
\bibinfo{author}{Everingham, M.}, \bibinfo{author}{Van~Gool, L.},
  \bibinfo{author}{Williams, C.K.}, \bibinfo{author}{Winn, J.},
  \bibinfo{author}{Zisserman, A.}, \bibinfo{year}{2010}.
\newblock \bibinfo{title}{The pascal visual object classes (voc) challenge}.
\newblock \bibinfo{journal}{International journal of computer vision}
  \bibinfo{volume}{88}, \bibinfo{pages}{303--338}.
\bibitem[{Feng et~al.(2020)Feng, Zhou, Cheng, Tan, Shi and Ma}]{feng2020semiv1}
\bibinfo{author}{Feng, Z.}, \bibinfo{author}{Zhou, Q.}, \bibinfo{author}{Cheng,
  G.}, \bibinfo{author}{Tan, X.}, \bibinfo{author}{Shi, J.},
  \bibinfo{author}{Ma, L.}, \bibinfo{year}{2020}.
\newblock \bibinfo{title}{Semi-supervised semantic segmentation via dynamic
  self-training and class-balanced curriculum}.
\newblock \bibinfo{journal}{arXiv preprint arXiv:2004.08514v1} .
\bibitem[{Fernando et~al.(2013)Fernando, Habrard, Sebban and
  Tuytelaars}]{sub_space}
\bibinfo{author}{Fernando, B.}, \bibinfo{author}{Habrard, A.},
  \bibinfo{author}{Sebban, M.}, \bibinfo{author}{Tuytelaars, T.},
  \bibinfo{year}{2013}.
\newblock \bibinfo{title}{Unsupervised visual domain adaptation using subspace
  alignment}, in: \bibinfo{booktitle}{ICCV}.
\bibitem[{French et~al.(2020a)French, Laine, Aila, Laine, Mackiewicz and
  Finlayson}]{french2019semi}
\bibinfo{author}{French, G.}, \bibinfo{author}{Laine, S.},
  \bibinfo{author}{Aila, T.}, \bibinfo{author}{Laine, S.},
  \bibinfo{author}{Mackiewicz, M.}, \bibinfo{author}{Finlayson, G.},
  \bibinfo{year}{2020}a.
\newblock \bibinfo{title}{Semi-supervised semantic segmentation needs strong,
  varied perturbations}, in: \bibinfo{booktitle}{British Machine Vision
  Conference}.
\bibitem[{French et~al.(2018)French, Mackiewicz and Fisher}]{french2018self}
\bibinfo{author}{French, G.}, \bibinfo{author}{Mackiewicz, M.},
  \bibinfo{author}{Fisher, M.}, \bibinfo{year}{2018}.
\newblock \bibinfo{title}{Self-ensembling for visual domain adaptation}, in:
  \bibinfo{booktitle}{Proceedings of the International Conference on Learning
  Representations}.
\bibitem[{French et~al.(2020b)French, Oliver and Salimans}]{french2020milking}
\bibinfo{author}{French, G.}, \bibinfo{author}{Oliver, A.},
  \bibinfo{author}{Salimans, T.}, \bibinfo{year}{2020}b.
\newblock \bibinfo{title}{Milking cowmask for semi-supervised image
  classification}.
\newblock \bibinfo{journal}{arXiv preprint arXiv:2003.12022} .
\bibitem[{Gaidon et~al.(2016a)Gaidon, Wang, Cabon and Vig}]{kITTI}
\bibinfo{author}{Gaidon, A.}, \bibinfo{author}{Wang, Q.},
  \bibinfo{author}{Cabon, Y.}, \bibinfo{author}{Vig, E.},
  \bibinfo{year}{2016}a.
\newblock \bibinfo{title}{Virtual worlds as proxy for multi-object tracking
  analysis}, in: \bibinfo{booktitle}{CVPR}.
\bibitem[{Gaidon et~al.(2016b)Gaidon, Wang, Cabon and Vig}]{VKITTI}
\bibinfo{author}{Gaidon, A.}, \bibinfo{author}{Wang, Q.},
  \bibinfo{author}{Cabon, Y.}, \bibinfo{author}{Vig, E.},
  \bibinfo{year}{2016}b.
\newblock \bibinfo{title}{Virtual worlds as proxy for multi-object tracking
  analysis}, in: \bibinfo{booktitle}{CVPR}.
\bibitem[{Ganin and Lempitsky(2015)}]{DANN}
\bibinfo{author}{Ganin, Y.}, \bibinfo{author}{Lempitsky, V.},
  \bibinfo{year}{2015}.
\newblock \bibinfo{title}{Unsupervised domain adaptation by backpropagation},
  in: \bibinfo{booktitle}{ICLR}.
\bibitem[{Geng et~al.(2011)Geng, Tao and Xu}]{MMD}
\bibinfo{author}{Geng, B.}, \bibinfo{author}{Tao, D.}, \bibinfo{author}{Xu,
  C.}, \bibinfo{year}{2011}.
\newblock \bibinfo{title}{Daml: Domain adaptation metric learning}.
\newblock \bibinfo{journal}{TIP} \bibinfo{volume}{20},
  \bibinfo{pages}{2980--2989}.
\bibitem[{Gong et~al.(2012)Gong, Shi, Sha and Grauman}]{flow_kernel}
\bibinfo{author}{Gong, B.}, \bibinfo{author}{Shi, Y.}, \bibinfo{author}{Sha,
  F.}, \bibinfo{author}{Grauman, K.}, \bibinfo{year}{2012}.
\newblock \bibinfo{title}{Geodesic flow kernel for unsupervised domain
  adaptation}, in: \bibinfo{booktitle}{CVPR}.
\bibitem[{Gong et~al.(2019)Gong, Li, Chen and Gool}]{DLOW}
\bibinfo{author}{Gong, R.}, \bibinfo{author}{Li, W.}, \bibinfo{author}{Chen,
  Y.}, \bibinfo{author}{Gool, L.V.}, \bibinfo{year}{2019}.
\newblock \bibinfo{title}{Dlow: Domain flow for adaptation and generalization},
  in: \bibinfo{booktitle}{CVPR}.
\bibitem[{Goodfellow et~al.(2014)Goodfellow, Pouget-Abadie, Mirza, Xu,
  Warde-Farley, Ozair, Courville and Bengio}]{GAN}
\bibinfo{author}{Goodfellow, I.}, \bibinfo{author}{Pouget-Abadie, J.},
  \bibinfo{author}{Mirza, M.}, \bibinfo{author}{Xu, B.},
  \bibinfo{author}{Warde-Farley, D.}, \bibinfo{author}{Ozair, S.},
  \bibinfo{author}{Courville, A.}, \bibinfo{author}{Bengio, Y.},
  \bibinfo{year}{2014}.
\newblock \bibinfo{title}{Generative adversarial nets}.
\newblock \bibinfo{journal}{Advances in neural information processing systems}
  \bibinfo{volume}{27}.
\bibitem[{Gu et~al.(2021)Gu, Zhou, Xu, Feng, Cheng, Lu, Shi and Ma}]{PIT}
\bibinfo{author}{Gu, Q.}, \bibinfo{author}{Zhou, Q.}, \bibinfo{author}{Xu, M.},
  \bibinfo{author}{Feng, Z.}, \bibinfo{author}{Cheng, G.}, \bibinfo{author}{Lu,
  X.}, \bibinfo{author}{Shi, J.}, \bibinfo{author}{Ma, L.},
  \bibinfo{year}{2021}.
\newblock \bibinfo{title}{Pit: Position-invariant transform for cross-fov
  domain adaptation}, in: \bibinfo{booktitle}{Proceedings of the IEEE/CVF
  International Conference on Computer Vision}.
\bibitem[{Han et~al.(2019)Han, Zou, Gao, Wang and
  Metaxas}]{han2019unsupervised}
\bibinfo{author}{Han, L.}, \bibinfo{author}{Zou, Y.}, \bibinfo{author}{Gao,
  R.}, \bibinfo{author}{Wang, L.}, \bibinfo{author}{Metaxas, D.},
  \bibinfo{year}{2019}.
\newblock \bibinfo{title}{Unsupervised domain adaptation via calibrating
  uncertainties}, in: \bibinfo{booktitle}{CVPR Workshops}.
\bibitem[{He et~al.(2016)He, Zhang, Ren and Sun}]{he2016deep}
\bibinfo{author}{He, K.}, \bibinfo{author}{Zhang, X.}, \bibinfo{author}{Ren,
  S.}, \bibinfo{author}{Sun, J.}, \bibinfo{year}{2016}.
\newblock \bibinfo{title}{Deep residual learning for image recognition}, in:
  \bibinfo{booktitle}{Proceedings of the IEEE conference on computer vision and
  pattern recognition}, pp. \bibinfo{pages}{770--778}.
\bibitem[{Hoffman et~al.(2018)Hoffman, Tzeng, Park, Zhu, Isola, Saenko, Efros
  and Darrell}]{Cycada}
\bibinfo{author}{Hoffman, J.}, \bibinfo{author}{Tzeng, E.},
  \bibinfo{author}{Park, T.}, \bibinfo{author}{Zhu, J.Y.},
  \bibinfo{author}{Isola, P.}, \bibinfo{author}{Saenko, K.},
  \bibinfo{author}{Efros, A.}, \bibinfo{author}{Darrell, T.},
  \bibinfo{year}{2018}.
\newblock \bibinfo{title}{{C}y{CADA}: Cycle-consistent adversarial domain
  adaptation}, in: \bibinfo{booktitle}{International conference on machine
  learning}, pp. \bibinfo{pages}{1989--1998}.
\bibitem[{Hoffman et~al.(2016)Hoffman, Wang, Yu and Darrell}]{FCN_wild}
\bibinfo{author}{Hoffman, J.}, \bibinfo{author}{Wang, D.}, \bibinfo{author}{Yu,
  F.}, \bibinfo{author}{Darrell, T.}, \bibinfo{year}{2016}.
\newblock \bibinfo{title}{Fcns in the wild: Pixel-level adversarial and
  constraint-based adaptation}.
\newblock \bibinfo{journal}{CoRR} \bibinfo{volume}{abs/1612.02649}.
\bibitem[{Huang et~al.(2020)Huang, Lu, Guan and Zhang}]{CrCDA}
\bibinfo{author}{Huang, J.}, \bibinfo{author}{Lu, S.}, \bibinfo{author}{Guan,
  D.}, \bibinfo{author}{Zhang, X.}, \bibinfo{year}{2020}.
\newblock \bibinfo{title}{Contextual-relation consistent domain adaptation for
  semantic segmentation}, in: \bibinfo{booktitle}{European conference on
  computer vision}, pp. \bibinfo{pages}{705--722}.
\bibitem[{Kendall and Gal(2017)}]{kendall2017uncertainties}
\bibinfo{author}{Kendall, A.}, \bibinfo{author}{Gal, Y.}, \bibinfo{year}{2017}.
\newblock \bibinfo{title}{What uncertainties do we need in bayesian deep
  learning for computer vision?}, in: \bibinfo{booktitle}{NeurIPS}.
\bibitem[{Kim and Byun(2020)}]{LTIR}
\bibinfo{author}{Kim, M.}, \bibinfo{author}{Byun, H.}, \bibinfo{year}{2020}.
\newblock \bibinfo{title}{Learning texture invariant representation for domain
  adaptation of semantic segmentation}, in: \bibinfo{booktitle}{Proc. CVPR},
  pp. \bibinfo{pages}{12975--12984}.
\bibitem[{Kulis et~al.(2011)Kulis, Saenko and Darrell}]{asymmetric_kernel}
\bibinfo{author}{Kulis, B.}, \bibinfo{author}{Saenko, K.},
  \bibinfo{author}{Darrell, T.}, \bibinfo{year}{2011}.
\newblock \bibinfo{title}{What you saw is not what you get: Domain adaptation
  using asymmetric kernel transforms}, in: \bibinfo{booktitle}{CVPR}.
\bibitem[{Kurmi et~al.(2019)Kurmi, Kumar and Namboodiri}]{kurmi2019attending}
\bibinfo{author}{Kurmi, V.K.}, \bibinfo{author}{Kumar, S.},
  \bibinfo{author}{Namboodiri, V.P.}, \bibinfo{year}{2019}.
\newblock \bibinfo{title}{Attending to discriminative certainty for domain
  adaptation}, in: \bibinfo{booktitle}{Proceedings of the IEEE/CVF Conference
  on Computer Vision and Pattern Recognition}, pp. \bibinfo{pages}{491--500}.
\bibitem[{Laine and Aila(2016)}]{laine2016temporal}
\bibinfo{author}{Laine, S.}, \bibinfo{author}{Aila, T.}, \bibinfo{year}{2016}.
\newblock \bibinfo{title}{Temporal ensembling for semi-supervised learning}.
\newblock \bibinfo{journal}{arXiv preprint arXiv:1610.02242} .
\bibitem[{LeCun et~al.(1998)LeCun, Bottou, Bengio and Haffner}]{MNIST}
\bibinfo{author}{LeCun, Y.}, \bibinfo{author}{Bottou, L.},
  \bibinfo{author}{Bengio, Y.}, \bibinfo{author}{Haffner, P.},
  \bibinfo{year}{1998}.
\newblock \bibinfo{title}{Gradient-based learning applied to document
  recognition}.
\newblock \bibinfo{journal}{IEEE Proc.} \bibinfo{volume}{86},
  \bibinfo{pages}{2278--2324}.
\bibitem[{Lee et~al.(2019a)Lee, Batra, Baig and Ulbricht}]{SWD}
\bibinfo{author}{Lee, C.Y.}, \bibinfo{author}{Batra, T.},
  \bibinfo{author}{Baig, M.H.}, \bibinfo{author}{Ulbricht, D.},
  \bibinfo{year}{2019}a.
\newblock \bibinfo{title}{Sliced wasserstein discrepancy for unsupervised
  domain adaptation}, in: \bibinfo{booktitle}{CVPR}.
\bibitem[{Lee et~al.(2019b)Lee, Ros, Li and Gaidon}]{lee2018spigan}
\bibinfo{author}{Lee, K.H.}, \bibinfo{author}{Ros, G.}, \bibinfo{author}{Li,
  J.}, \bibinfo{author}{Gaidon, A.}, \bibinfo{year}{2019}b.
\newblock \bibinfo{title}{Spigan: Privileged adversarial learning from
  simulation} .
\bibitem[{Li et~al.(2020a)Li, Kang, Liu, Wei and Yang}]{CCM}
\bibinfo{author}{Li, G.}, \bibinfo{author}{Kang, G.}, \bibinfo{author}{Liu,
  W.}, \bibinfo{author}{Wei, Y.}, \bibinfo{author}{Yang, Y.},
  \bibinfo{year}{2020}a.
\newblock \bibinfo{title}{Content-consistent matching for domain adaptive
  semantic segmentation}, in: \bibinfo{booktitle}{European conference on
  computer vision}, \bibinfo{organization}{Springer}. pp.
  \bibinfo{pages}{440--456}.
\bibitem[{Li et~al.(2020b)Li, Liu, Lin, Xie, Ding, Huang and Tang}]{DCAN}
\bibinfo{author}{Li, S.}, \bibinfo{author}{Liu, H.C.}, \bibinfo{author}{Lin,
  Q.}, \bibinfo{author}{Xie, B.}, \bibinfo{author}{Ding, Z.},
  \bibinfo{author}{Huang, G.}, \bibinfo{author}{Tang, J.},
  \bibinfo{year}{2020}b.
\newblock \bibinfo{title}{Domain conditioned adaptation network}, in:
  \bibinfo{booktitle}{Proc. AAAI}, pp. \bibinfo{pages}{11386--11393}.
\bibitem[{Li et~al.(2019a)Li, Zhong, Wu, Yang, Lin and Liu}]{EMANet}
\bibinfo{author}{Li, X.}, \bibinfo{author}{Zhong, Z.}, \bibinfo{author}{Wu,
  J.}, \bibinfo{author}{Yang, Y.}, \bibinfo{author}{Lin, Z.},
  \bibinfo{author}{Liu, H.}, \bibinfo{year}{2019}a.
\newblock \bibinfo{title}{Expectation-maximization attention networks for
  semantic segmentation}, in: \bibinfo{booktitle}{ICCV}.
\bibitem[{Li et~al.(2019b)Li, Yuan and Vasconcelos}]{BDL}
\bibinfo{author}{Li, Y.}, \bibinfo{author}{Yuan, L.},
  \bibinfo{author}{Vasconcelos, N.}, \bibinfo{year}{2019}b.
\newblock \bibinfo{title}{Bidirectional learning for domain adaptation of
  semantic segmentation}, in: \bibinfo{booktitle}{Proceedings of the IEEE/CVF
  Conference on Computer Vision and Pattern Recognition}, pp.
  \bibinfo{pages}{6936--6945}.
\bibitem[{Lian et~al.(2019)Lian, Duan, Lv and Gong}]{PyCDA}
\bibinfo{author}{Lian, Q.}, \bibinfo{author}{Duan, L.}, \bibinfo{author}{Lv,
  F.}, \bibinfo{author}{Gong, B.}, \bibinfo{year}{2019}.
\newblock \bibinfo{title}{Constructing self-motivated pyramid curriculums for
  cross-domain semantic segmentation: {A} non-adversarial approach}, in:
  \bibinfo{booktitle}{Proceedings of the IEEE/CVF International Conference on
  Computer Vision}, pp. \bibinfo{pages}{6757--6766}.
\bibitem[{Lin et~al.(2014)Lin, Maire, Belongie, Hays, Perona, Ramanan,
  Doll{\'a}r and Zitnick}]{COCO}
\bibinfo{author}{Lin, T.Y.}, \bibinfo{author}{Maire, M.},
  \bibinfo{author}{Belongie, S.}, \bibinfo{author}{Hays, J.},
  \bibinfo{author}{Perona, P.}, \bibinfo{author}{Ramanan, D.},
  \bibinfo{author}{Doll{\'a}r, P.}, \bibinfo{author}{Zitnick, C.L.},
  \bibinfo{year}{2014}.
\newblock \bibinfo{title}{Microsoft coco: Common objects in context}, in:
  \bibinfo{booktitle}{European conference on computer vision},
  \bibinfo{organization}{Springer}. pp. \bibinfo{pages}{740--755}.
\bibitem[{Long et~al.(2015a)Long, Shelhamer and Darrell}]{fcn}
\bibinfo{author}{Long, J.}, \bibinfo{author}{Shelhamer, E.},
  \bibinfo{author}{Darrell, T.}, \bibinfo{year}{2015}a.
\newblock \bibinfo{title}{Fully convolutional networks for semantic
  segmentation}, in: \bibinfo{booktitle}{CVPR}.
\bibitem[{Long et~al.(2015b)Long, Cao, Wang and Jordan}]{long2015learning}
\bibinfo{author}{Long, M.}, \bibinfo{author}{Cao, Y.}, \bibinfo{author}{Wang,
  J.}, \bibinfo{author}{Jordan, M.}, \bibinfo{year}{2015}b.
\newblock \bibinfo{title}{Learning transferable features with deep adaptation
  networks}, in: \bibinfo{booktitle}{ICLR}.
\bibitem[{Lu et~al.(2020)Lu, Yang, Zhu, Liu, Song and Xiang}]{STAR}
\bibinfo{author}{Lu, Z.}, \bibinfo{author}{Yang, Y.}, \bibinfo{author}{Zhu,
  X.}, \bibinfo{author}{Liu, C.}, \bibinfo{author}{Song, Y.Z.},
  \bibinfo{author}{Xiang, T.}, \bibinfo{year}{2020}.
\newblock \bibinfo{title}{Stochastic classifiers for unsupervised domain
  adaptation}, in: \bibinfo{booktitle}{Proceedings of the IEEE/CVF Conference
  on Computer Vision and Pattern Recognition}, pp. \bibinfo{pages}{9111--9120}.
\bibitem[{Luo et~al.(2019a)Luo, Liu, Guan, Yu and Yang}]{SIBAN}
\bibinfo{author}{Luo, Y.}, \bibinfo{author}{Liu, P.}, \bibinfo{author}{Guan,
  T.}, \bibinfo{author}{Yu, J.}, \bibinfo{author}{Yang, Y.},
  \bibinfo{year}{2019}a.
\newblock \bibinfo{title}{Significance-aware information bottleneck for domain
  adaptive semantic segmentation}, in: \bibinfo{booktitle}{Proceedings of the
  IEEE/CVF International Conference on Computer Vision}, pp.
  \bibinfo{pages}{6778--6787}.
\bibitem[{Luo et~al.(2021)Luo, Liu, Zheng, Guan, Yu and Yang}]{CLANv2}
\bibinfo{author}{Luo, Y.}, \bibinfo{author}{Liu, P.}, \bibinfo{author}{Zheng,
  L.}, \bibinfo{author}{Guan, T.}, \bibinfo{author}{Yu, J.},
  \bibinfo{author}{Yang, Y.}, \bibinfo{year}{2021}.
\newblock \bibinfo{title}{Category-level adversarial adaptation for semantic
  segmentation using purified features}.
\newblock \bibinfo{journal}{IEEE Transactions on Pattern Analysis and Machine
  Intelligence} ,
  \bibinfo{pages}{1--1}\DOIprefix\doi{10.1109/TPAMI.2021.3064379}.
\bibitem[{Luo et~al.(2019b)Luo, Zheng, Guan, Yu and Yang}]{CLAN}
\bibinfo{author}{Luo, Y.}, \bibinfo{author}{Zheng, L.}, \bibinfo{author}{Guan,
  T.}, \bibinfo{author}{Yu, J.}, \bibinfo{author}{Yang, Y.},
  \bibinfo{year}{2019}b.
\newblock \bibinfo{title}{Taking a closer look at domain shift: Category-level
  adversaries for semantics consistent domain adaptation}, in:
  \bibinfo{booktitle}{Proceedings of the IEEE/CVF Conference on Computer Vision
  and Pattern Recognition}, pp. \bibinfo{pages}{2507--2516}.
\bibitem[{{Naseer Subhani} and {Ali}(2020)}]{LSE}
\bibinfo{author}{{Naseer Subhani}, M.}, \bibinfo{author}{{Ali}, M.},
  \bibinfo{year}{2020}.
\newblock \bibinfo{title}{Learning from scale-invariant examples for domain
  adaptation in semantic segmentation}, in: \bibinfo{booktitle}{European
  conference on computer vision}, \bibinfo{organization}{Springer}. pp.
  \bibinfo{pages}{290--306}.
\bibitem[{Netzer et~al.(2011)Netzer, Wang, Coates, Bissacco, Wu and Ng}]{SVHN}
\bibinfo{author}{Netzer, Y.}, \bibinfo{author}{Wang, T.},
  \bibinfo{author}{Coates, A.}, \bibinfo{author}{Bissacco, A.},
  \bibinfo{author}{Wu, B.}, \bibinfo{author}{Ng, A.Y.}, \bibinfo{year}{2011}.
\newblock \bibinfo{title}{Reading digits in natural images with unsupervised
  feature learning}, in: \bibinfo{booktitle}{NeurIPS workshop}.
\bibitem[{Pan et~al.(2020)Pan, Shin, Rameau, Lee and Kweon}]{IntraDA}
\bibinfo{author}{Pan, F.}, \bibinfo{author}{Shin, I.}, \bibinfo{author}{Rameau,
  F.}, \bibinfo{author}{Lee, S.}, \bibinfo{author}{Kweon, I.S.},
  \bibinfo{year}{2020}.
\newblock \bibinfo{title}{Unsupervised intra-domain adaptation for semantic
  segmentation through self-supervision}, in: \bibinfo{booktitle}{Unsupervised
  Intra-domain Adaptation for Semantic Segmentation through Self-Supervision},
  pp. \bibinfo{pages}{3764--3773}.
\bibitem[{Paul et~al.(2020)Paul, Tsai, Schulter, Roy{-}Chowdhury and
  Chandraker}]{WLabel}
\bibinfo{author}{Paul, S.}, \bibinfo{author}{Tsai, Y.},
  \bibinfo{author}{Schulter, S.}, \bibinfo{author}{Roy{-}Chowdhury, A.K.},
  \bibinfo{author}{Chandraker, M.}, \bibinfo{year}{2020}.
\newblock \bibinfo{title}{Domain adaptive semantic segmentation using weak
  labels}, in: \bibinfo{booktitle}{European conference on computer vision},
  \bibinfo{organization}{Springer}. pp. \bibinfo{pages}{571--587}.
\bibitem[{Perone et~al.(2019)Perone, Ballester, Barros and
  Cohen-Adad}]{medical_self_emsembling}
\bibinfo{author}{Perone, C.S.}, \bibinfo{author}{Ballester, P.},
  \bibinfo{author}{Barros, R.C.}, \bibinfo{author}{Cohen-Adad, J.},
  \bibinfo{year}{2019}.
\newblock \bibinfo{title}{Unsupervised domain adaptation for medical imaging
  segmentation with self-ensembling}.
\newblock \bibinfo{journal}{NeuroImage} \bibinfo{volume}{194},
  \bibinfo{pages}{1--11}.
\bibitem[{Richter et~al.(2016)Richter, Vineet, Roth and Koltun}]{gtav}
\bibinfo{author}{Richter, S.R.}, \bibinfo{author}{Vineet, V.},
  \bibinfo{author}{Roth, S.}, \bibinfo{author}{Koltun, V.},
  \bibinfo{year}{2016}.
\newblock \bibinfo{title}{Playing for data: Ground truth from computer games},
  in: \bibinfo{booktitle}{ECCV}.
\bibitem[{Ros et~al.(2016)Ros, Sellart, Materzynska, Vazquez and
  Lopez}]{synthia}
\bibinfo{author}{Ros, G.}, \bibinfo{author}{Sellart, L.},
  \bibinfo{author}{Materzynska, J.}, \bibinfo{author}{Vazquez, D.},
  \bibinfo{author}{Lopez, A.M.}, \bibinfo{year}{2016}.
\newblock \bibinfo{title}{The synthia dataset: A large collection of synthetic
  images for semantic segmentation of urban scenes}, in:
  \bibinfo{booktitle}{CVPR}.
\bibitem[{Saito et~al.(2018)Saito, Watanabe, Ushiku and Harada}]{mcd}
\bibinfo{author}{Saito, K.}, \bibinfo{author}{Watanabe, K.},
  \bibinfo{author}{Ushiku, Y.}, \bibinfo{author}{Harada, T.},
  \bibinfo{year}{2018}.
\newblock \bibinfo{title}{Maximum classifier discrepancy for unsupervised
  domain adaptation}, in: \bibinfo{booktitle}{CVPR}.
\bibitem[{Sankaranarayanan et~al.(2018)Sankaranarayanan, Balaji, Jain, Nam~Lim
  and Chellappa}]{LSD}
\bibinfo{author}{Sankaranarayanan, S.}, \bibinfo{author}{Balaji, Y.},
  \bibinfo{author}{Jain, A.}, \bibinfo{author}{Nam~Lim, S.},
  \bibinfo{author}{Chellappa, R.}, \bibinfo{year}{2018}.
\newblock \bibinfo{title}{Learning from synthetic data: Addressing domain shift
  for semantic segmentation}, in: \bibinfo{booktitle}{CVPR}.
\bibitem[{Simonyan and Andrew(2014)}]{VGG}
\bibinfo{author}{Simonyan, K.}, \bibinfo{author}{Andrew, Z.},
  \bibinfo{year}{2014}.
\newblock \bibinfo{title}{Very deep convolutional networks for large-scale
  image recognition}.
\newblock \bibinfo{journal}{arXiv preprint arXiv:1409.1556} .
\bibitem[{Sun et~al.(2016)Sun, Feng and Saenko}]{sun2016return}
\bibinfo{author}{Sun, B.}, \bibinfo{author}{Feng, J.}, \bibinfo{author}{Saenko,
  K.}, \bibinfo{year}{2016}.
\newblock \bibinfo{title}{Return of frustratingly easy domain adaptation}, in:
  \bibinfo{booktitle}{AAAI}.
\bibitem[{Tarvainen and Valpola(2017)}]{Mean_teacher}
\bibinfo{author}{Tarvainen, A.}, \bibinfo{author}{Valpola, H.},
  \bibinfo{year}{2017}.
\newblock \bibinfo{title}{Mean teachers are better role models: Weight-averaged
  consistency targets improve semi-supervised deep learning results}, in:
  \bibinfo{booktitle}{Advances in Neural Information Processing Systems 30},
  pp. \bibinfo{pages}{1195--1204}.
\bibitem[{Tranheden et~al.(2021)Tranheden, Olsson, Pinto and
  Svensson}]{tranheden2020dacs}
\bibinfo{author}{Tranheden, W.}, \bibinfo{author}{Olsson, V.},
  \bibinfo{author}{Pinto, J.}, \bibinfo{author}{Svensson, L.},
  \bibinfo{year}{2021}.
\newblock \bibinfo{title}{Dacs: Domain adaptation via cross-domain mixed
  sampling}, in: \bibinfo{booktitle}{Proceedings of the IEEE/CVF Winter
  Conference on Applications of Computer Vision}, pp.
  \bibinfo{pages}{1379--1389}.
\bibitem[{Tsai et~al.(2018)Tsai, Hung, Schulter, Sohn, Yang and
  Chandraker}]{AdaptSegNet}
\bibinfo{author}{Tsai, Y.H.}, \bibinfo{author}{Hung, W.C.},
  \bibinfo{author}{Schulter, S.}, \bibinfo{author}{Sohn, K.},
  \bibinfo{author}{Yang, M.H.}, \bibinfo{author}{Chandraker, M.},
  \bibinfo{year}{2018}.
\newblock \bibinfo{title}{Learning to adapt structured output space for
  semantic segmentation}, in: \bibinfo{booktitle}{Proceedings of the IEEE
  conference on computer vision and pattern recognition}, pp.
  \bibinfo{pages}{7472--7481}.
\bibitem[{Tsai et~al.(2019)Tsai, Sohn, Schulter and
  Chandraker}]{tsai2019domain}
\bibinfo{author}{Tsai, Y.H.}, \bibinfo{author}{Sohn, K.},
  \bibinfo{author}{Schulter, S.}, \bibinfo{author}{Chandraker, M.},
  \bibinfo{year}{2019}.
\newblock \bibinfo{title}{Domain adaptation for structured output via
  discriminative patch representations}, in: \bibinfo{booktitle}{Proceedings of
  the IEEE/CVF International Conference on Computer Vision}, pp.
  \bibinfo{pages}{1456--1465}.
\bibitem[{Vu et~al.(2019a)Vu, Jain, Bucher, Cord and P{\'{e}}rez}]{DADA}
\bibinfo{author}{Vu, T.}, \bibinfo{author}{Jain, H.}, \bibinfo{author}{Bucher,
  M.}, \bibinfo{author}{Cord, M.}, \bibinfo{author}{P{\'{e}}rez, P.},
  \bibinfo{year}{2019}a.
\newblock \bibinfo{title}{{DADA:} depth-aware domain adaptation in semantic
  segmentation}, in: \bibinfo{booktitle}{Proceedings of the IEEE/CVF
  International Conference on Computer Vision}, pp.
  \bibinfo{pages}{7363--7372}.
\bibitem[{Vu et~al.(2019b)Vu, Jain, Bucher, Cord and P{\'e}rez}]{advent}
\bibinfo{author}{Vu, T.H.}, \bibinfo{author}{Jain, H.},
  \bibinfo{author}{Bucher, M.}, \bibinfo{author}{Cord, M.},
  \bibinfo{author}{P{\'e}rez, P.}, \bibinfo{year}{2019}b.
\newblock \bibinfo{title}{Advent: Adversarial entropy minimization for domain
  adaptation in semantic segmentation}, in: \bibinfo{booktitle}{CVPR}.
\bibitem[{Wang et~al.(2020a)Wang, Shen, Zhang, Duan and Mei}]{FADA}
\bibinfo{author}{Wang, H.}, \bibinfo{author}{Shen, T.}, \bibinfo{author}{Zhang,
  W.}, \bibinfo{author}{Duan, L.}, \bibinfo{author}{Mei, T.},
  \bibinfo{year}{2020}a.
\newblock \bibinfo{title}{Classes matter: A fine-grained adversarial approach
  to cross-domain semantic segmentation}, \bibinfo{organization}{Springer}. pp.
  \bibinfo{pages}{642--659}.
\bibitem[{Wang et~al.(2020b)Wang, Yu, Wei, Feris, Xiong, Hwu, Huang and
  Shi}]{SIM}
\bibinfo{author}{Wang, Z.}, \bibinfo{author}{Yu, M.}, \bibinfo{author}{Wei,
  Y.}, \bibinfo{author}{Feris, R.}, \bibinfo{author}{Xiong, J.},
  \bibinfo{author}{Hwu, W.m.}, \bibinfo{author}{Huang, T.S.},
  \bibinfo{author}{Shi, H.}, \bibinfo{year}{2020}b.
\newblock \bibinfo{title}{Differential treatment for stuff and things: A simple
  unsupervised domain adaptation method for semantic segmentation}, in:
  \bibinfo{booktitle}{Proceedings of the IEEE/CVF Conference on Computer Vision
  and Pattern Recognition}, pp. \bibinfo{pages}{12635--12644}.
\bibitem[{Wen et~al.(2019)Wen, Zheng, Yuan, Gong and Chen}]{wen2019bayesian}
\bibinfo{author}{Wen, J.}, \bibinfo{author}{Zheng, N.}, \bibinfo{author}{Yuan,
  J.}, \bibinfo{author}{Gong, Z.}, \bibinfo{author}{Chen, C.},
  \bibinfo{year}{2019}.
\newblock \bibinfo{title}{Bayesian uncertainty matching for unsupervised domain
  adaptation}.
\newblock \bibinfo{journal}{arXiv preprint arXiv:1906.09693} .
\bibitem[{Xu et~al.(2019)Xu, Du, Zhang, Zhang, Wang and Zhang}]{SEANet}
\bibinfo{author}{Xu, Y.}, \bibinfo{author}{Du, B.}, \bibinfo{author}{Zhang,
  L.}, \bibinfo{author}{Zhang, Q.}, \bibinfo{author}{Wang, G.},
  \bibinfo{author}{Zhang, L.}, \bibinfo{year}{2019}.
\newblock \bibinfo{title}{Self-ensembling attention networks: Addressing domain
  shift for semantic segmentation}, in: \bibinfo{booktitle}{Proceedings of the
  AAAI Conference on Artificial Intelligence}, pp. \bibinfo{pages}{5581--5588}.
\bibitem[{Yang et~al.(2020a)Yang, An, Wang, Zhu, Yan and Huang}]{LDR}
\bibinfo{author}{Yang, J.}, \bibinfo{author}{An, W.}, \bibinfo{author}{Wang,
  S.}, \bibinfo{author}{Zhu, X.}, \bibinfo{author}{Yan, C.},
  \bibinfo{author}{Huang, J.}, \bibinfo{year}{2020}a.
\newblock \bibinfo{title}{Label-driven reconstruction for domain adaptation in
  semantic segmentation}, in: \bibinfo{booktitle}{European conference on
  computer vision}, \bibinfo{organization}{Springer}. pp.
  \bibinfo{pages}{480--498}.
\bibitem[{Yang et~al.(2021)Yang, An, Yan, Zhao and Huang}]{yang2021context}
\bibinfo{author}{Yang, J.}, \bibinfo{author}{An, W.}, \bibinfo{author}{Yan,
  C.}, \bibinfo{author}{Zhao, P.}, \bibinfo{author}{Huang, J.},
  \bibinfo{year}{2021}.
\newblock \bibinfo{title}{Context-aware domain adaptation in semantic
  segmentation}, in: \bibinfo{booktitle}{Proceedings of the IEEE/CVF Winter
  Conference on Applications of Computer Vision}, pp.
  \bibinfo{pages}{514--524}.
\bibitem[{Yang et~al.(2020b)Yang, Xu, Li, Qi, Shen, Li and Lin}]{APODA}
\bibinfo{author}{Yang, J.}, \bibinfo{author}{Xu, R.}, \bibinfo{author}{Li, R.},
  \bibinfo{author}{Qi, X.}, \bibinfo{author}{Shen, X.}, \bibinfo{author}{Li,
  G.}, \bibinfo{author}{Lin, L.}, \bibinfo{year}{2020}b.
\newblock \bibinfo{title}{An adversarial perturbation oriented domain
  adaptation approach for semantic segmentation}, in:
  \bibinfo{booktitle}{Proceedings of the AAAI Conference on Artificial
  Intelligence}, pp. \bibinfo{pages}{12613--12620}.
\bibitem[{Yang et~al.(2020c)Yang, Lao, Sundaramoorthi and Soatto}]{PCEDA}
\bibinfo{author}{Yang, Y.}, \bibinfo{author}{Lao, D.},
  \bibinfo{author}{Sundaramoorthi, G.}, \bibinfo{author}{Soatto, S.},
  \bibinfo{year}{2020}c.
\newblock \bibinfo{title}{Phase consistent ecological domain adaptation}, in:
  \bibinfo{booktitle}{Proceedings of the IEEE/CVF Conference on Computer Vision
  and Pattern Recognition}, pp. \bibinfo{pages}{9011--9020}.
\bibitem[{Yang and Soatto(2020)}]{FDA}
\bibinfo{author}{Yang, Y.}, \bibinfo{author}{Soatto, S.}, \bibinfo{year}{2020}.
\newblock \bibinfo{title}{Fda: Fourier domain adaptation for semantic
  segmentation}, in: \bibinfo{booktitle}{Proceedings of the IEEE/CVF Conference
  on Computer Vision and Pattern Recognition}, pp. \bibinfo{pages}{4085--4095}.
\bibitem[{Yu et~al.(2021)Yu, Zhang, Dong, Hu, Dong and Zhang}]{DAST}
\bibinfo{author}{Yu, F.}, \bibinfo{author}{Zhang, M.}, \bibinfo{author}{Dong,
  H.}, \bibinfo{author}{Hu, S.}, \bibinfo{author}{Dong, B.},
  \bibinfo{author}{Zhang, L.}, \bibinfo{year}{2021}.
\newblock \bibinfo{title}{Dast: Unsupervised domain adaptation in semantic
  segmentation based on discriminator attention and self-training}, in:
  \bibinfo{booktitle}{Proceedings of the AAAI Conference on Artificial
  Intelligence}, pp. \bibinfo{pages}{10754--10762}.
\bibitem[{Yu et~al.(2019)Yu, Wang, Li, Fu and Heng}]{yu2019uncertainty}
\bibinfo{author}{Yu, L.}, \bibinfo{author}{Wang, S.}, \bibinfo{author}{Li, X.},
  \bibinfo{author}{Fu, C.W.}, \bibinfo{author}{Heng, P.A.},
  \bibinfo{year}{2019}.
\newblock \bibinfo{title}{Uncertainty-aware self-ensembling model for
  semi-supervised 3d left atrium segmentation}, in:
  \bibinfo{booktitle}{MICCAI}.
\bibitem[{Zhang et~al.(2019)Zhang, Zhang, Liu and Tao}]{CAG_UDA}
\bibinfo{author}{Zhang, Q.}, \bibinfo{author}{Zhang, J.}, \bibinfo{author}{Liu,
  W.}, \bibinfo{author}{Tao, D.}, \bibinfo{year}{2019}.
\newblock \bibinfo{title}{Category anchor-guided unsupervised domain adaptation
  for semantic segmentation}, in: \bibinfo{booktitle}{Advances in Neural
  Information Processing Systems}, pp. \bibinfo{pages}{433--443}.
\bibitem[{Zhang et~al.(2017)Zhang, David and Gong}]{CDA}
\bibinfo{author}{Zhang, Y.}, \bibinfo{author}{David, P.},
  \bibinfo{author}{Gong, B.}, \bibinfo{year}{2017}.
\newblock \bibinfo{title}{Curriculum domain adaptation for semantic
  segmentation of urban scenes}, in: \bibinfo{booktitle}{Proceedings of the
  IEEE international conference on computer vision}, pp.
  \bibinfo{pages}{2020--2030}.
\bibitem[{Zhao et~al.(2019)Zhao, Li, Yue, Gu, Xu, Tan, Chai and
  Keutzer}]{MADAN_NIPS2019}
\bibinfo{author}{Zhao, S.}, \bibinfo{author}{Li, B.}, \bibinfo{author}{Yue,
  X.}, \bibinfo{author}{Gu, Y.}, \bibinfo{author}{Xu, P.},
  \bibinfo{author}{Tan, Hu, R.}, \bibinfo{author}{Chai, H.},
  \bibinfo{author}{Keutzer, K.}, \bibinfo{year}{2019}.
\newblock \bibinfo{title}{Multi-source domain adaptation for semantic
  segmentation}, in: \bibinfo{booktitle}{NeurIPS}.
\bibitem[{Zheng and Yang(2020)}]{zheng2020unsupervised}
\bibinfo{author}{Zheng, Z.}, \bibinfo{author}{Yang, Y.}, \bibinfo{year}{2020}.
\newblock \bibinfo{title}{Rectifying pseudo label learning via uncertainty
  estimation for domain adaptive semantic segmentation}.
\newblock \bibinfo{journal}{International Journal of Computer Vision (IJCV)}
  \DOIprefix\doi{10.1007/s11263-020-01395-y}.
\bibitem[{Zhou et~al.(2020)Zhou, Wang, Chu, Yang, Bai and Xu}]{ASA}
\bibinfo{author}{Zhou, W.}, \bibinfo{author}{Wang, Y.}, \bibinfo{author}{Chu,
  J.}, \bibinfo{author}{Yang, J.}, \bibinfo{author}{Bai, X.},
  \bibinfo{author}{Xu, Y.}, \bibinfo{year}{2020}.
\newblock \bibinfo{title}{Affinity space adaptation for semantic segmentation
  across domains}.
\newblock \bibinfo{journal}{IEEE Transactions on Image Processing}
  \bibinfo{volume}{30}, \bibinfo{pages}{2549--2561}.
\bibitem[{Zhu et~al.(2017)Zhu, Park, Isola and Efros}]{CycleGAN2017}
\bibinfo{author}{Zhu, J.Y.}, \bibinfo{author}{Park, T.},
  \bibinfo{author}{Isola, P.}, \bibinfo{author}{Efros, A.A.},
  \bibinfo{year}{2017}.
\newblock \bibinfo{title}{Unpaired image-to-image translation using
  cycle-consistent adversarial networks}, in: \bibinfo{booktitle}{Proceedings
  of the IEEE international conference on computer vision}, pp.
  \bibinfo{pages}{2223--2232}.
\bibitem[{Zhu et~al.(2018)Zhu, Zhou, Yang, Shi and Lin}]{Conservative_loss}
\bibinfo{author}{Zhu, X.}, \bibinfo{author}{Zhou, H.}, \bibinfo{author}{Yang,
  C.}, \bibinfo{author}{Shi, J.}, \bibinfo{author}{Lin, D.},
  \bibinfo{year}{2018}.
\newblock \bibinfo{title}{Penalizing top performers: Conservative loss for
  semantic segmentation adaptation}, in: \bibinfo{booktitle}{Proceedings of the
  European Conference on Computer Vision (ECCV)}, pp.
  \bibinfo{pages}{568--583}.
\bibitem[{Zou et~al.(2018)Zou, Yu, Kumar and Wang}]{CBST}
\bibinfo{author}{Zou, Y.}, \bibinfo{author}{Yu, Z.}, \bibinfo{author}{Kumar,
  B.}, \bibinfo{author}{Wang, J.}, \bibinfo{year}{2018}.
\newblock \bibinfo{title}{Unsupervised domain adaptation for semantic
  segmentation via class-balanced self-training}, in:
  \bibinfo{booktitle}{Proceedings of the European conference on computer vision
  (ECCV)}, pp. \bibinfo{pages}{289--305}.
\bibitem[{Zou et~al.(2019)Zou, Yu, Liu, Kumar and Wang}]{CRST}
\bibinfo{author}{Zou, Y.}, \bibinfo{author}{Yu, Z.}, \bibinfo{author}{Liu, X.},
  \bibinfo{author}{Kumar, B.}, \bibinfo{author}{Wang, J.},
  \bibinfo{year}{2019}.
\newblock \bibinfo{title}{Confidence regularized self-training}, in:
  \bibinfo{booktitle}{Proceedings of the IEEE/CVF International Conference on
  Computer Vision}, pp. \bibinfo{pages}{5982--5991}.

\end{thebibliography}

\end{document}